\documentclass{article}

\PassOptionsToPackage{numbers, sort&compress}{natbib}

\usepackage[main, final]{neurips_2026}

\usepackage[utf8]{inputenc}
\usepackage[T1]{fontenc}
\usepackage{hyperref}
\usepackage{url}
\usepackage{booktabs}
\usepackage{amsfonts}
\usepackage{nicefrac}
\usepackage{microtype}
\usepackage{graphicx}
\usepackage{amsmath}
\usepackage{multirow}
\usepackage{pifont}
\usepackage{amssymb}
\usepackage[table,xcdraw]{xcolor}
\definecolor{qualgreen}{HTML}{00B050}
\definecolor{qualred}{HTML}{FF0000}
\usepackage{wrapfig}
\usepackage{enumitem}

\title{Before It Fades: Reinforcing Temporal Representations at Inference Time in VideoLLMs}

\author{%
  Youngwoo Shin \qquad Yusung Ro \qquad Minseo Kim \qquad Junmo Kim \\[8pt]
  \mdseries Korea Advanced Institute of Science and Technology (KAIST) \\[8pt]
  \mdseries \texttt{\{yshin0917, ysro3067, alstj1571, junmo.kim\}@kaist.ac.kr}
}

\begin{document}

\maketitle

\vspace{-5pt}

\begin{abstract}
Video Large Language Models (VideoLLMs) receive frames in sequential order and interpret how visual content evolves along the temporal axis, yet temporal reasoning remains a persistent weakness across architectures. Reversing the frame order of a video, a transformation that should invert temporal answers, often leaves the final prediction unchanged. We investigate where this failure originates by defining the temporal divergence vector $\tau_l$, the layer-wise representational difference induced by reversing temporal order. Tracking its magnitude across layers reveals a consistent temporal divergence profile where the divergence peaks at intermediate layers and progressively diminishes toward the output. We confirm this peak is specific to temporal reasoning and functionally critical for predictions, establishing that VideoLLMs acquire temporal information at intermediate layers but fail to maintain it to the output. This progressive fading motivates our method, Temporal Activation Injection (TAI), which extracts $\tau_l$ at the peak of the profile for each input and reinjects it into subsequent layers following the measured decay. TAI requires no training and consistently improves temporal reasoning across three VideoLLMs and four benchmarks with negligible impact on non-temporal tasks. Code is available at \url{https://github.com/Youngwoo-git/Before-It-Fades}.
\end{abstract}

\vspace{-8pt}

\section{Introduction}
\label{sec:intro}

\begin{figure*}[htbp]
    \centering
    \includegraphics[width=0.97\textwidth]{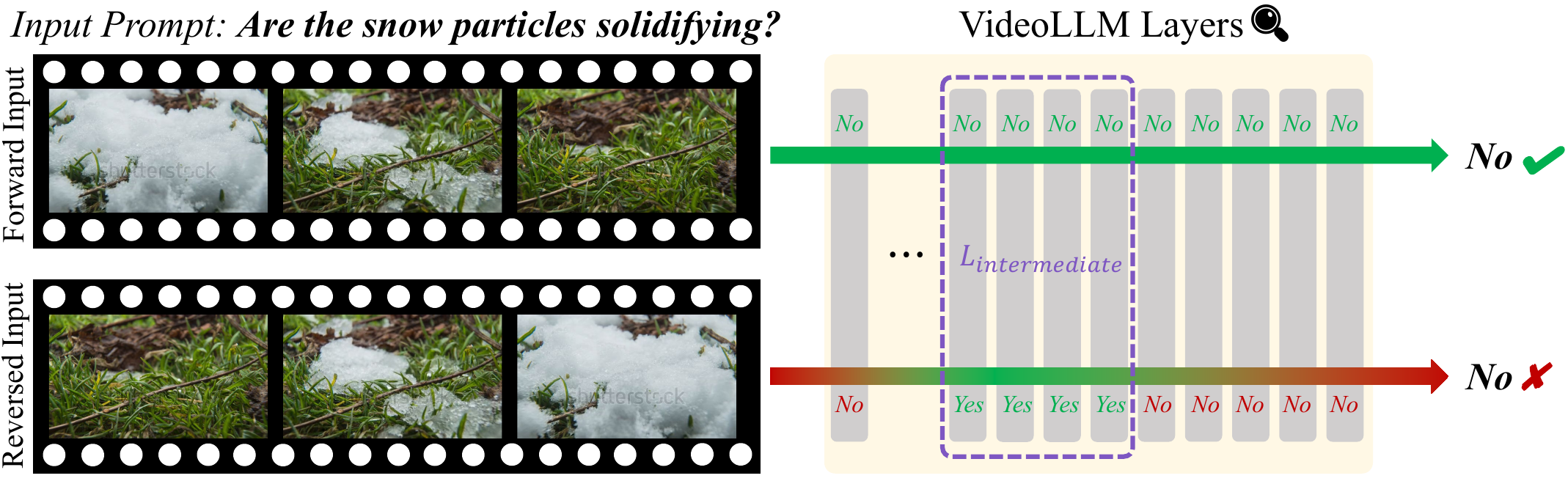}
    \caption{\textbf{A representative failure case of temporal 
    reasoning.} Logit lens~\citep{nostalgebraist2020logitlens} shows the 
    higher-probability answer between Yes and No at each 
    layer. The forward video maintains the correct answer 
    throughout, while the reversed video shifts toward the 
    correct answer at $L_{\mathrm{intermediate}}$ before 
    subsequent layers overturn it, converging both inputs to 
    the same prediction by $L_{\mathrm{last}}$.}
    \label{fig:problem_statement}
\end{figure*}

Video Large Language Models (VideoLLMs) enable 
interpreting ordered sequences of visual frames through 
the reasoning capabilities of large language 
models, advancing video understanding across tasks~\citep{bai2025qwen25vltechnicalreport, bai2025qwen3, 
chen2024internvl}. Temporal reasoning is central to this 
process, as these models receive frames in sequential 
order and interpret how visual content evolves along the 
temporal axis. Yet it remains a persistent weakness 
across architectures~\citep{cores2024lost, 
liu2024tempcompass, xue2025seeing}, and 
Fig.~\ref{fig:problem_statement} illustrates a 
representative failure case. Reversing the frame order of 
a video is a transformation that should invert the answer 
to a temporal question, yet the model produces the same 
prediction for both inputs, failing to perform temporal 
reasoning.

Improving temporal reasoning in VideoLLMs has been 
approached from multiple 
directions. Training-based methods learn temporal 
capabilities through dedicated 
objectives~\citep{xue2025seeing, li2025temporal}. Inference-time methods, predominantly contrastive 
decoding~\citep{zhang2024eventhallusion, qi2025improve,
wu2025seasonmitigatingtemporalhallucination}, contrast 
output distributions from temporally distorted inputs to 
adjust final predictions. Both regard temporal reasoning 
failure as a given limitation and address it externally, 
without investigating where it originates. 
However, we observe in Fig.~\ref{fig:problem_statement} 
that this failure may not be inherent, as the 
representation at $L_{\mathrm{intermediate}}$ assigns a 
higher probability to the correct answer for the reversed 
input before subsequent layers overturn it. This 
indicates that representations sensitive to temporal 
order emerge but are not maintained through the remaining 
layers, raising our central question: \emph{where do representations sensitive to temporal order concentrate and shift across layers, and how can we intervene there to improve temporal reasoning?}

To answer this question, we quantify how these 
representations change across layers, identifying where 
they concentrate and how they evolve toward the output. 
To isolate the contribution of temporal order in these 
representations, we reverse the frame order of each video 
to construct a pair that differs only along the temporal 
axis while sharing all spatial content and text 
conditioning. To measure this contribution at each layer, 
we define the representational difference between the two 
inputs as the \textit{temporal divergence vector} 
$\tau_l$. Tracking its normalized magnitude across layers 
produces the \textit{temporal divergence profile}, which 
follows a shared pattern across all architectures we 
evaluate: the divergence peaks in the mid-to-late layers, 
then progressively diminishes toward the 
output. We confirm that this peak is specific to temporal 
reasoning through question-conditioned analysis and that 
it is functionally critical for the model's predictions 
through causal intervention. These results establish that 
VideoLLMs acquire temporal information at intermediate 
layers and rely on it for their predictions, yet this 
information progressively fades before reaching the 
output.

This progressive fading presents a natural intervention 
point, motivating our method, \textbf{Temporal Activation 
Injection} (TAI), which extracts $\tau_l$ at the peak of 
the temporal divergence profile for each input and 
reinjects it into subsequent layers following the 
measured decay. $\tau_l$ captures the direction and 
magnitude by which temporal information is expressed in 
the representation at each layer, and adding it to 
subsequent layers compensates for the signal that the 
model produces but fails to maintain. TAI 
uses the temporal divergence profile itself as layer-wise 
weights, grounding the intervention in quantities 
measured from the model. TAI requires no training, adapts 
to each input through its own $\tau_l$, and operates 
entirely at inference time without modifying model 
weights.

We evaluate TAI across three VideoLLMs with diverse 
architectures~\citep{bai2025qwen25vltechnicalreport, 
bai2025qwen3, chen2024internvl} on temporal reasoning 
benchmarks~\citep{liu2024tempcompass, cores2024lost, 
xue2025seeing} and a general video understanding 
benchmark~\citep{li2024mvbench}. TAI consistently 
improves temporal reasoning across all three models with 
negligible impact on non-temporal performance. 
Improvements concentrate on categories whose answers 
depend on frame order, while categories invariant under 
reversal receive minimal effect, confirming that TAI 
selectively targets temporal reasoning. Across all 
experiments, the temporal divergence profile serves as 
both an analytical tool that reveals where temporal 
information concentrates and a practical basis for 
targeted intervention.

Our contributions are as follows:
\begin{itemize}[leftmargin=*, itemsep=2pt, parsep=0pt]
    \item We introduce the temporal divergence vector 
    $\tau_l$ and its layer-wise profile, revealing that 
    VideoLLMs acquire temporal information at 
    intermediate layers but fail to maintain it to the 
    output.
    \item We propose Temporal Activation Injection (TAI), 
    a training-free method that leverages the temporal 
    divergence profile to improve temporal reasoning in 
    VideoLLMs.
    \item We demonstrate consistent improvements in 
    temporal reasoning across three VideoLLMs and four 
    benchmarks with negligible impact on non-temporal 
    tasks.
\end{itemize}

\section{Related Works}

\subsection{Temporal Reasoning in VideoLLMs}
VideoLLMs~\citep{bai2025qwen25vltechnicalreport, bai2025qwen3,
chen2024internvl, li2024llava, lin2024video, zhang2024llava}
integrate video encoders~\citep{zhao2024videoprism,
wang2024internvideo2} with LLM backbones and adopt
temporal-aware enhancements~\citep{shi2026causality,
rasekh2025enhancing}, yet consistently struggle with temporal
reasoning~\citep{cores2024lost, liu2024tempcompass}.
Prior work has approached this limitation from multiple angles.
Through probing and training, T3~\citep{li2025temporal}
demonstrates that video embeddings capture sufficient temporal
information yet the LLM decoder fails to leverage it,
localizing the bottleneck to the language model without
identifying where within it the deficiency arises.
ArrowRL~\citep{xue2025seeing} observes that models show little
sensitivity at the output level when videos are reversed or
shuffled, and addresses it through reinforcement learning.
Mechanistic analyses offer a complementary perspective.
Kim et al.~\citep{kim2026map} find that temporal reasoning
relies on a sparse set of effective pathways rather than the
full attention graph, and causal interventions on the last
token confirm that temporal information is progressively
aggregated into the prediction position through cross-modal
attention~\citep{shi2026causality}. However, these analyses
trace how temporal information flows but not how this signal evolves in subsequent layers.
Our work builds on these findings, establishing that the 
temporal signal peaks at intermediate layers and 
progressively diminishes toward the output, and addresses 
this decay without additional training.

\subsection{Inference-Time Model Steering}
Recent methods steer LLM behavior at inference time by
manipulating either intermediate hidden states or output
distributions, and several have been adapted to improve
temporal faithfulness in VideoLLMs. In the first direction,
Representation Engineering
(RepE)~\citep{zou2023representation} controls model behavior
by manipulating concept directions within hidden states
without modifying model weights.
ITI~\citep{li2023inferencetime} trains linear probes per
attention head to locate truthful directions, and
CAA~\citep{rimsky2024steering} computes mean activation
differences from contrastive pairs.
SADI~\citep{wang2025semanticsadaptive} and
CAST~\citep{lee2025programming} adapt the intervention to
input context but still rely on labeled data to learn the
steering directions. Rather than steering hidden states, a
separate line of work targets temporal faithfulness through
contrastive decoding. TCD~\citep{zhang2024eventhallusion}, 
VTD~\citep{qi2025improve}, and 
SEASON~\citep{wu2025seasonmitigatingtemporalhallucination} 
contrast output distributions from temporally distorted 
inputs to suppress unfaithful predictions, while DINO-HEAL~\citep{li2025vidhalluc} adjusts 
predictions using features from a separate vision model. These methods operate 
without examining the model's internal representations, 
intervening at the output level or relying on external 
signals. Our analysis reveals that temporal signals peak 
at intermediate layers and substantially degrade by the 
time logits are produced. Our method operates at the 
intermediate representation level, extracting 
input-specific temporal vectors where temporal 
information concentrates without ground-truth annotations or parameter updates.

\section{Analysis of Temporal Representations}
\label{sec:analysis}

Reversing the temporal order of frames alters the internal 
representations of models during inference, but where and 
how much the activations diverge across layers has not 
been characterized. We term this layer-wise 
activation divergence the \textit{temporal divergence}. Since the reversed video shares identical spatial content 
and text conditioning with the original, any difference 
in hidden states isolates the contribution of temporal 
order. We 
pair each input video $V$ with its time-reversed 
counterpart $\tilde{V}$, obtained by reversing the frame 
sequence along the temporal axis. We conduct our analysis across three VideoLLMs of 
comparable scale with diverse architectures: 
Qwen2.5-VL-7B~\citep{bai2025qwen25vltechnicalreport}, Qwen3-VL-8B~\citep{bai2025qwen3}, and InternVL2.5-8B~\citep{chen2024internvl}. We use 50 temporally reversed video pairs 
with binary temporal questions from 
TempCompass~\citep{liu2024tempcompass}, where ground-truth 
answers invert under reversal.

\begin{figure*}[!t]
    \centering
    \includegraphics[width=0.98\textwidth]{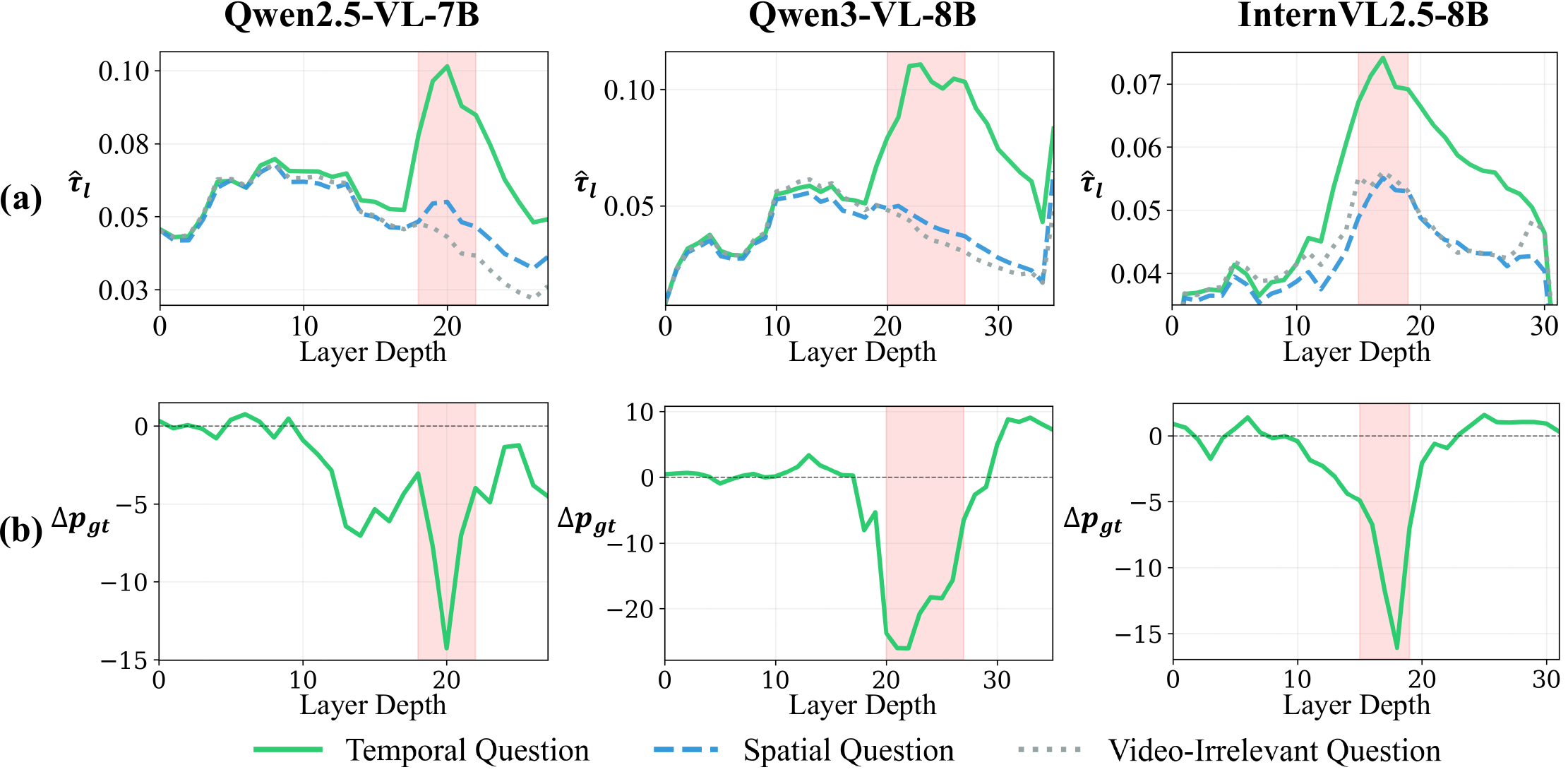}
    \caption{\textbf{Layer-wise temporal divergence analysis 
    across three VideoLLMs.} \textbf{(a)}~$\hat{\tau}_l$ 
    measured on frame-order-reversed pairs under temporal, 
    spatial, and video-irrelevant questions. The temporal 
    divergence profile, peaking in the mid-to-late 
    layers and diminishing toward the output, emerges 
    only under temporal questions, while both non-temporal 
    conditions remain substantially attenuated. 
    \textbf{(b)}~Change in ground-truth probability 
    $p_{\mathrm{gt}}$ in \%p when the last token is blocked from 
    attending to preceding positions per layer. The 
    largest drop aligns with the $\hat{\tau}_l$ peak, 
    confirming these layers are functionally critical for 
    temporal predictions. Red shaded regions mark the peak 
    layers.}
    \label{fig:analysis}
\end{figure*}

\subsection{Layer-wise Temporal Divergence}
\label{sec:tau_def}

To track temporal divergence across layers, we focus on 
the last token, which under causal attention is the only 
position that has attended to all preceding visual and 
textual tokens~\citep{zou2023representation, 
li2023inferencetime}. We extract the hidden state $h_l 
\in \mathbb{R}^{d}$ at this position for each layer $l 
\in \{0, \ldots, L{-}1\}$, where $L$ is the number 
of layers and $d$ the hidden dimension. We denote the 
last token hidden states under the two inputs as 
$h_l^{\mathrm{fwd}}$ and $h_l^{\mathrm{rev}}$, and define 
the \textit{temporal divergence vector} $\tau_l = 
h_l^{\mathrm{fwd}} - h_l^{\mathrm{rev}}$. The magnitude of $\tau_l$ at a given layer reflects how 
much the hidden state changes when frame 
order is reversed. For fair comparison across layers, we normalize 
by the forward activation norm to account for the 
monotonic growth of hidden state magnitudes with layer 
depth, yielding the scalar \textit{normalized temporal 
divergence} $\hat{\tau}_l$:

\begin{equation}
    \hat{\tau}_l = \frac{\|\tau_l\|}{\|h_l^{\mathrm{fwd}}\|} 
    = \frac{\|h_l^{\mathrm{fwd}} - h_l^{\mathrm{rev}}\|}
    {\|h_l^{\mathrm{fwd}}\|}.
    \label{eq:tau_norm}
\end{equation}

We compute $\hat{\tau}_l$ for each of the three models and 
report the mean temporal divergence profiles in 
Fig.~\ref{fig:analysis}~(a, solid) using temporal 
questions from TempCompass~\citep{liu2024tempcompass}. A consistent pattern emerges across all architectures. Through the initial layers, $\hat{\tau}_l$ remains low as the model processes visual and textual tokens without substantial divergence. Starting from the intermediate layers, $\hat{\tau}_l$ rises sharply and peaks in the mid-to-late layers. After this peak, $\hat{\tau}_l$ progressively declines 
toward the final layers, falling to values comparable to 
or below those at the start, though some models exhibit a 
slight uptick near the last layer consistent with hidden 
state jumps~\citep{shibata-etal-2026-suppressing}. The convergence of this profile across models with different vision encoders, LLM backbones, positional encoding schemes, and training recipes indicates that $\hat{\tau}_l$ captures a shared computational regularity in how decoder-only VideoLLMs process frame order dependent information. 

This convergence across architectures reflects more than 
the flow of temporal information through the model, as 
$\hat{\tau}_l$ quantifies where temporal information 
concentrates at the representation level across layers. Reversing 
the frame sequence inverts the answer to temporal 
questions while keeping the prompt and all other input 
components intact, so any divergence in the resulting 
representations isolates the effect of altering the 
temporal axis. By tracking this divergence 
at the last token, the position whose representation 
drives the final prediction, $\hat{\tau}_l$ captures how 
temporal information reaches the prediction across 
layers. Because $\hat{\tau}_l$ is computed only with 
hidden states from a single forward pass on each video, 
it requires no ground-truth annotations, trained probes, or model 
modification, and can be applied to characterize any 
VideoLLM that exposes intermediate representations.

\subsection{Temporal Specificity of Divergence}
\label{sec:specificity}

$\hat{\tau}_l$ has favorable properties as a diagnostic for finding the place of divergence, yet whether the divergence at the peak is bound to temporal reasoning specifically remains to be verified. The peak in $\hat{\tau}_l$ may capture every representational difference caused by reversal, or it may reflect the subset specific to temporal reasoning. To examine which of these holds, we hold the video pair constant and vary only the question. If the divergence at the peak is determined by the video alone, $\hat{\tau}_l$ should remain stable across different questions on the same video pair. If instead $\hat{\tau}_l$ responds to what the model is asked to reason about, the profile should shift with the question. We test this by evaluating $\hat{\tau}_l$ on the same forward and reversed video pairs under three question conditions. (1) Temporal questions from TempCompass~\citep{liu2024tempcompass} that require frame order awareness to answer (e.g., \textit{Is the person moving downwards?}), and two control conditions we construct to isolate temporal specificity: (2) a spatial question whose answer does not change under reversal (\textit{Is this video filmed indoors or outdoors?}), and (3) a video-irrelevant question unrelated to visual content (\textit{Does a week have seven days?}).

Fig.~\ref{fig:analysis}~(a, dashed and dotted) illustrates
$\hat{\tau}_l$ under the two non-temporal conditions on
the same video pairs. Unlike the temporal condition
described in Sec.~\ref{sec:tau_def}, the distinctive peak
is substantially attenuated and $\hat{\tau}_l$ remains
comparatively low through the later layers. The same
video pairs produce different $\hat{\tau}_l$ profiles
depending solely on the question, confirming that the
peak reflects the last token selectively aggregating
temporal information when the task demands it rather than
passively inheriting frame order differences from the
visual input. The peak in $\hat{\tau}_l$ therefore provides a reliable, question-conditioned indicator of where to intervene for temporal reasoning.

\subsection{Functional Importance and Signal Decay}
\label{sec:knockout}

The peak in $\hat{\tau}_l$ emerges where the last token selectively aggregates temporal information, but whether the model actually depends on this signal for its predictions remains to be tested. To assess this, we apply attention knockout~\citep{geva2023dissecting} to the last token at each layer, setting the attention mask $M_l(t, s) = -\infty$ for the last token position $t$ attending to all positions $s$ at layer $l$. This removes the last token's access to the video and prompt tokens it would normally attend to, blocking the flow of information into the prediction position at that layer. We sweep this intervention across all layers and measure the resulting change in ground truth answer probability $p_{\mathrm{gt}}$ on subtasks related to temporal reasoning~\citep{liu2024tempcompass}. Fig.~\ref{fig:analysis} (b) shows that blocking layers in the early or final stages produces modest effects, while blocking the region around the $\hat{\tau}_l$ peak causes the largest probability drop. This region aligns precisely with the $\hat{\tau}_l$ peak in Fig.~\ref{fig:analysis} (a), confirming that the temporal information concentrated at these layers is actively utilized for the final prediction rather than an incidental activation pattern. The same layers that produce the largest temporal divergence in Sec.~\ref{sec:tau_def} and respond selectively to temporal questions in Sec.~\ref{sec:specificity} are also where the last token draws most heavily on temporal information for its prediction.

The $\hat{\tau}_l$ profiles in Fig.~\ref{fig:analysis} (a) further reveal that this temporal information does not persist beyond the peak. After the peak, $\hat{\tau}_l$ declines progressively across all three models, consistent with broader findings that later LLM layers contribute diminishing semantic value~\citep{gromov2025the}. This decay at the representation level accounts for the prediction-level failure observed in Fig.~\ref{fig:problem_statement}, where the correct answer at $L_{\mathrm{intermediate}}$ is overturned by subsequent layers. The profile therefore identifies both where to secure the temporal information before it fades and how much compensation each subsequent layer requires, presenting a natural intervention point grounded in the temporal divergence profile of the model. The experiments in Sec.~\ref{sec:experiments} validate that compensating for this decay consistently improves temporal reasoning across all models, with the specificity analysis in Sec.~\ref{sec:steering} confirming that the compensation selectively targets temporal categories of the benchmarks with minimal effect on non-temporal ones, suggesting the decay constitutes an addressable stage of inference for temporal reasoning.

\section{Temporal Activation Injection}
\label{sec:tai}

 \begin{figure*}[t]
    \centering
    \includegraphics[width=0.96\textwidth]{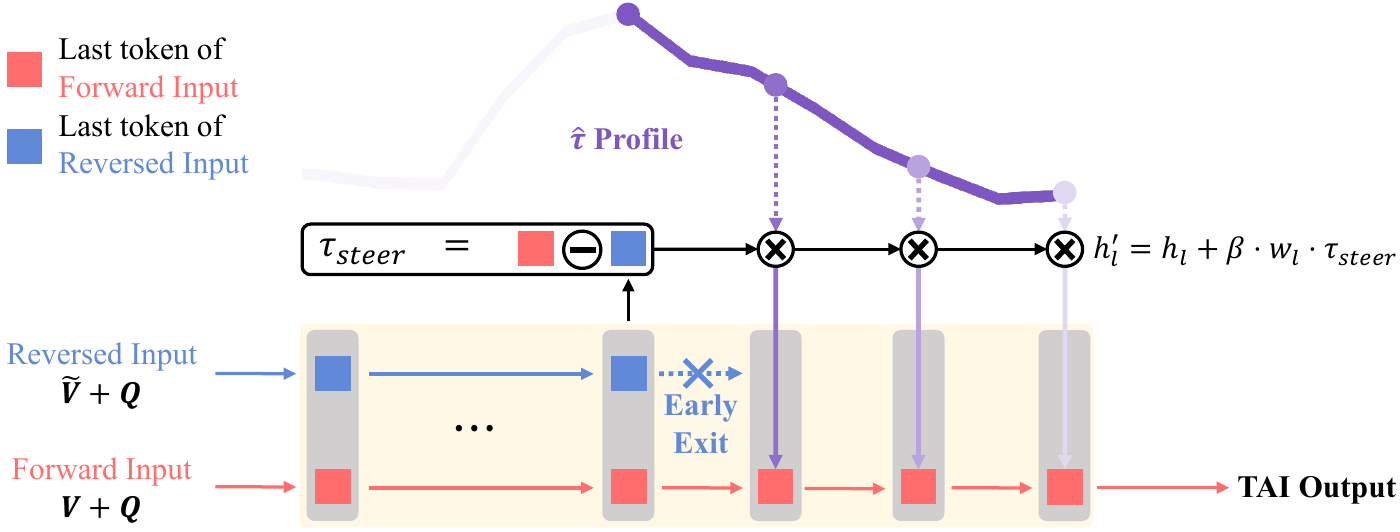}
    \caption{\textbf{Overview of Temporal Activation Injection (TAI).} Given a video $V$, its time-reversed counterpart $\tilde{V}$, and question prompt $Q$, the reversed input is processed up to $L_{\mathrm{src}}$ and early-exited, bypassing all subsequent layers. $\tau_{\mathrm{steer}}$ is computed at $L_{\mathrm{src}}$ from the difference between forward and reversed last-token hidden states. The $\hat{\tau}$ profile (top) provides the per-layer weights $w_l$ that scale the injection into each subsequent layer of the forward pass, producing the TAI output.}
    \label{fig:pipeline}
\end{figure*}

The natural intervention point identified in Sec.~\ref{sec:analysis}, where temporal information concentrates and subsequently diminishes, motivates our method, Temporal Activation Injection (TAI), illustrated in Fig.~\ref{fig:pipeline}. TAI extracts temporal information at the peak of the $\hat{\tau}_l$ profile and reinjects it into the last token at subsequent layers where this information diminishes, compensating for the decay identified in Sec.~\ref{sec:knockout}. The entire process operates on hidden states that the model already computes during inference, requiring no ground-truth annotations, additional training, or model modification.

\noindent\textbf{Extraction.} Given an input video $V$ and 
prompt $Q$, we construct its time-reversed 
counterpart $\tilde{V}$ by reversing the frame sequence. 
The analysis in Sec.~\ref{sec:analysis} identified that 
temporal information concentrates around the peak layers 
of the $\hat{\tau}_l$ profile and progressively decays 
afterward. We therefore set the source layer 
$L_{\mathrm{src}}$ to the peak immediately before this 
decay begins and process $\tilde{V}$ through the model up 
to $L_{\mathrm{src}}$, capturing the last token hidden 
state 
$h_{L_{\mathrm{src}}}^{\mathrm{rev}} \in \mathbb{R}^{d}$. 
All layers beyond $L_{\mathrm{src}}$ are bypassed via 
early termination to avoid unnecessary computation. We 
then process $V$ through the full model, capturing 
$h_{L_{\mathrm{src}}}^{\mathrm{fwd}}$ at the same layer 
and applying injection at subsequent layers as described 
below.
\begin{equation}
    \tau_{\mathrm{steer}} \triangleq \tau_{L_{\mathrm{src}}} = h_{L_{\mathrm{src}}}^{\mathrm{fwd}} - h_{L_{\mathrm{src}}}^{\mathrm{rev}}.
    \label{eq:tau_extract}
\end{equation}

We do not normalize $\tau_{\mathrm{steer}}$ to utilize the 
full directional vector in the representation space. 
While the profile-derived weights $w_l$ distribute the 
injection across layers at the model level, the 
magnitude of $\tau_{\mathrm{steer}}$ adapts the injection 
strength at the input level, so that videos sensitive to 
frame reversal naturally receive stronger injection while 
temporally invariant inputs receive near-zero correction. 
Unlike fixed steering vectors, $\tau_{\mathrm{steer}}$ is 
extracted separately for each video and remains 
input-specific.

\noindent\textbf{Profile-guided injection.} The $\hat{\tau}_l$ profile measured in Sec.~\ref{sec:tau_def} demonstrates that the temporal divergence concentrates around the peak layer and progressively decreases in the subsequent layers. We leverage this empirically measured profile directly as a per-layer injection schedule. For each layer $l > L_{\mathrm{src}}$, we modify the last token hidden state as:
\begin{equation}
    h_l' = h_l + \beta \cdot w_l \cdot \tau_{\mathrm{steer}}, \qquad
    w_l = \frac{\hat{\tau}_l}{\hat{\tau}_{L_{\mathrm{src}}}},
    \label{eq:injection}
\end{equation}

where $\beta$ is a global scaling coefficient and $w_l$ is pre-computed per model from the mean $\hat{\tau}_l$ profile, while $\tau_{\mathrm{steer}}$ adapts per input. Because $w_l$ follows the profile rather than a uniform or heuristic schedule, the injection is grounded in quantities measured from the model. Post-peak layers whose representations still carry a substantial temporal component receive proportionally larger injection, while deeper layers receive lighter additions that respect their established role in shaping the final prediction~\citep{gromov2025the, geva2023dissecting}. Since the $\hat{\tau}_l$ profile is measured at the last token position, injection is applied exclusively at that position, leaving all other token representations and model parameters unchanged.

\noindent\textbf{Implementation Details.} The source layer $L_{\mathrm{src}}$ is set to the peak of the $\hat{\tau}_l$ profile before the decay begins, which consistently falls in the mid-to-late layers across the models we evaluate. The profile weights $w_l$ are pre-computed once per model from 50 forward-reverse video pairs with temporal questions from TempCompass~\citep{liu2024tempcompass}, requiring only the videos and questions without any ground-truth annotations, and stored as a lightweight configuration. Since the profile captures a structural property of the model rather than properties of specific videos, it generalizes across inputs and stabilizes with relatively few samples, as detailed in Appendix~\ref{app:npairs}.

\section{Experiments}
\label{sec:experiments}

\subsection{Experimental Setup}
\label{sec:setup}
We apply TAI to three VideoLLMs of comparable scale with diverse architectures: Qwen2.5-VL-7B, Qwen3-VL-8B, and  InternVL2.5-8B. For temporal reasoning, we evaluate on TempCompass~\citep{liu2024tempcompass}, TVBench~\citep{cores2024lost}, and AoTBench~\citep{xue2025seeing}. For general video understanding, we use MVBench~\citep{li2024mvbench}. We compare against recent training-free methods TCD~\citep{zhang2024eventhallusion} and DINO-HEAL~\citep{li2025vidhalluc}, and the training-based ArrowRL~\citep{xue2025seeing}. All experiments are conducted on a single A6000 GPU with 16-frame uniform temporal sampling, and accuracy is reported in \%. The divergence profile required for profile-guided injection is computed once per model in Sec.~\ref{sec:analysis} and reused across all evaluations. The scaling coefficient $\beta$ in Eq.~\ref{eq:injection} is set per model and held constant across all benchmarks for that model.

\begin{table*}[htbp]
  \caption{\textbf{Temporal reasoning results by category on TempCompass and TVBench.} \textbf{Bold} and \underline{underline} mark the best and second best per group, and \colorbox{blue!10}{blue} highlights our method. All baselines are reproduced under our evaluation protocol, see Appendix~\ref{sec:reproduction}.}
  \label{tab:temporal-results}
  \centering
  \resizebox{\textwidth}{!}{%
  \begin{tabular}{lc cccccc cccccccccc}
    \toprule
    \multirow{2}{*}{\textbf{Models}} & \multirow{2}{*}{\textbf{\shortstack{Training-\\free}}}
    & \multicolumn{6}{c}{\textbf{TempCompass}} & \multicolumn{10}{c}{\textbf{TVBench}} \\
    \cmidrule(lr){3-8} \cmidrule(lr){9-18}
    & & Act & Attr & Dir & Ord & Spd & AVG & AC & AL & AS & ES & MD & OC & OS & ST & UA & AVG \\
    \midrule
    Qwen2.5-VL-7B & - & 94.9 & 77.2 & 59.1 & 75.8 & 59.2 & 73.4 & 25.9 & 37.5 & 62.9 & 42.0 & 32.3 & \underline{55.4} & \textbf{40.0} & 82.2 & 31.7 & 44.6 \\
    +TCD & \textcolor{green!60!black}{\checkmark} & 94.7 & 80.0 & 60.0 & \underline{77.6} & 58.8 & 74.2 & 26.1 & \textbf{40.6} & 63.8 & 42.5 & 33.6 & 54.1 & \underline{39.1} & 82.2 & 30.5 & 45.0 \\
    +DINO-HEAL & \textcolor{green!60!black}{\checkmark} & \textbf{95.5} & 74.9 & 58.2 & 74.8 & 59.0 & 72.7 & \underline{26.9} & \underline{40.0} & 64.1 & 42.0 & 31.5 & 53.4 & 38.7 & \underline{83.2} & \underline{32.9} & 45.0 \\
    +ArrowRL & \textcolor{red}{\ding{55}} & \underline{95.3} & \underline{82.8} & \underline{60.6} & 76.4 & \underline{59.7} & \underline{75.0} & \textbf{32.5} & 38.1 & \underline{66.8} & \textbf{43.5} & \textbf{34.9} & 54.7 & 31.6 & 80.5 & \textbf{46.3} & \textbf{46.9} \\
    \rowcolor{blue!10} \textbf{+TAI (Ours)} & \textcolor{green!60!black}{\checkmark} & 93.2 & \textbf{83.8} & \textbf{61.4} & \textbf{79.7} & \textbf{60.1} & \textbf{75.6} & 26.7 & \underline{40.0} & \textbf{67.0} & \underline{43.0} & \underline{34.5} & \textbf{58.8} & \underline{39.1} & \textbf{86.5} & 31.7 & \underline{46.6} \\
    \midrule
    Qwen3-VL-8B & - & \underline{94.7} & 79.7 & 60.5 & 84.1 & \textbf{66.6} & 77.1 & 28.4 & 40.0 & \underline{72.8} & 55.0 & \underline{53.4} & 43.9 & 33.3 & 81.6 & \textbf{51.2} & \underline{49.9} \\
    +TCD & \textcolor{green!60!black}{\checkmark} & 94.1 & \underline{80.8} & \underline{61.2} & \underline{84.7} & 65.5 & \underline{77.2} & \textbf{29.9} & 39.4 & 69.8 & \underline{57.5} & 51.3 & 44.6 & \underline{33.8} & \underline{83.8} & \textbf{51.2} & \underline{49.9} \\
    +DINO-HEAL & \textcolor{green!60!black}{\checkmark} & \textbf{95.1} & 80.4 & 60.3 & 84.1 & \underline{66.1} & \underline{77.2} & 28.5 & \textbf{41.2} & 71.4 & 53.5 & \underline{53.4} & \underline{45.3} & 31.6 & 83.2 & 48.8 & 49.6 \\
    \rowcolor{blue!10} \textbf{+TAI (Ours)} & \textcolor{green!60!black}{\checkmark} & 94.4 & \textbf{83.6} & \textbf{61.5} & \textbf{86.2} & 65.1 & \textbf{78.0} & \underline{29.1} & \underline{40.6} & \textbf{73.5} & \textbf{62.0} & \textbf{60.3} & \textbf{50.0} & \textbf{37.3} & \textbf{86.5} & 43.9 & \textbf{52.6} \\
    \midrule
    InternVL2.5-8B & - & \underline{91.8} & 74.1 & 54.0 & 74.9 & 59.3 & 70.9 & \underline{29.1} & 45.6 & \underline{65.4} & 75.0 & \textbf{97.4} & 62.2 & \textbf{38.7} & \textbf{87.6} & \underline{35.4} & 57.2 \\
    +TCD & \textcolor{green!60!black}{\checkmark} & 91.5 & \underline{78.8} & 54.0 & \underline{76.7} & 58.8 & \underline{72.0} & \textbf{31.3} & \textbf{46.2} & 63.8 & \textbf{76.5} & \textbf{97.4} & 62.2 & 36.4 & 86.5 & 34.1 & 57.2 \\
    +DINO-HEAL & \textcolor{green!60!black}{\checkmark} & 91.5 & 76.1 & \underline{54.9} & \underline{76.7} & \textbf{60.0} & 71.9 & \underline{29.1} & 45.6 & 63.8 & \underline{76.0} & 96.1 & \textbf{73.6} & 37.3 & \textbf{87.6} & 32.9 & \underline{57.4} \\
    \rowcolor{blue!10} \textbf{+TAI (Ours)} & \textcolor{green!60!black}{\checkmark} & \textbf{92.2} & \textbf{81.8} & \textbf{55.4} & \textbf{79.8} & \underline{59.4} & \textbf{73.6} & 28.9 & \textbf{46.2} & \textbf{70.5} & \underline{76.0} & 97.0 & \underline{63.5} & \underline{38.2} & 87.0 & \textbf{36.6} & \textbf{58.3} \\
    \bottomrule
  \end{tabular}%
  }
\end{table*}

\begin{table*}[htbp]
  \centering
  \begin{minipage}[t]{0.492\textwidth}
    \centering
    \caption{\textbf{Temporal reasoning results on AoTBench by subtask.} Notation as in Tab.~\ref{tab:temporal-results}.}
    \label{tab:aotbench-results}
    \small
    \resizebox{\textwidth}{!}{%
    \begin{tabular}{lc cccccc}
      \toprule
      \textbf{Models} & \textbf{\shortstack{Training-\\free}} & RevF & UCF & T2V & V2T & QA & AVG \\
      \midrule
      Qwen2.5-VL-7B & - & \underline{50.0} & \underline{51.6} & 53.8 & 63.2 & 46.7 & 54.5 \\
      +TCD & \textcolor{green!60!black}{\checkmark} & 49.3 & \underline{51.6} & 54.4 & 64.0 & 52.0 & 56.4 \\
      +DINO-HEAL & \textcolor{green!60!black}{\checkmark} & \underline{50.0} & 51.4 & \underline{54.9} & 63.9 & 50.0 & 55.9 \\
      +ArrowRL & \textcolor{red}{\ding{55}} & \textbf{50.7} & \textbf{55.2} & \textbf{55.1} & \underline{64.5} & \textbf{56.1} & \textbf{58.2} \\
      \rowcolor{blue!10} \textbf{+TAI (Ours)} & \textcolor{green!60!black}{\checkmark} & \underline{50.0} & \underline{51.6} & 53.8 & \textbf{65.9} & \underline{55.3} & \underline{57.7} \\
      \midrule
      Qwen3-VL-8B & - & \textbf{50.0} & \underline{51.6} & 51.3 & 62.2 & 58.1 & 56.6 \\
      +TCD & \textcolor{green!60!black}{\checkmark} & \textbf{50.0} & \textbf{51.8} & \textbf{51.6} & \underline{64.0} & \underline{60.6} & \underline{57.9} \\
      +DINO-HEAL & \textcolor{green!60!black}{\checkmark} & \textbf{50.0} & \underline{51.6} & \textbf{51.6} & 62.7 & 57.9 & 56.7 \\
      \rowcolor{blue!10} \textbf{+TAI (Ours)} & \textcolor{green!60!black}{\checkmark} & \textbf{50.0} & \underline{51.6} & 51.3 & \textbf{66.7} & \textbf{65.2} & \textbf{59.9} \\
      \midrule
      InternVL2.5-8B & - & \textbf{50.0} & \textbf{51.6} & 53.2 & 55.8 & 56.1 & 54.6 \\
      +TCD & \textcolor{green!60!black}{\checkmark} & \textbf{50.0} & \textbf{51.6} & 52.3 & \underline{57.2} & \underline{58.9} & \underline{55.6} \\
      +DINO-HEAL & \textcolor{green!60!black}{\checkmark} & \textbf{50.0} & \textbf{51.6} & \textbf{53.4} & 56.2 & 55.6 & 54.7 \\
      \rowcolor{blue!10} \textbf{+TAI (Ours)} & \textcolor{green!60!black}{\checkmark} & \textbf{50.0} & \textbf{51.6} & \textbf{53.4} & \textbf{58.2} & \textbf{60.1} & \textbf{56.6} \\
      \bottomrule
    \end{tabular}%
    }
  \end{minipage}%
  \hfill
  \begin{minipage}[t]{0.50\textwidth}
    \centering
    \caption{\textbf{General video understanding on MVBench.} Notation as in Tab.~\ref{tab:temporal-results}.}
    \label{tab:mvbench-results}
    \small
    \resizebox{\textwidth}{!}{%
    \begin{tabular}{lc cccc}
      \toprule
      \textbf{Models} & \textbf{\shortstack{Training-\\free}} & \textbf{\shortstack{Temporal-\\Relevant}} & \textbf{\shortstack{Temporal-\\Irrelevant}} & \textbf{Hybrid} & \textbf{AVG} \\
      \midrule
      Qwen2.5-VL-7B & - & 58.8 & \underline{58.5} & 77.7 & 62.9 \\
      +TCD & \textcolor{green!60!black}{\checkmark} & \underline{59.2} & 58.0 & 77.7 & 63.0 \\
      +DINO-HEAL & \textcolor{green!60!black}{\checkmark} & 58.7 & \textbf{58.6} & \underline{78.3} & \underline{63.1} \\
      +ArrowRL & \textcolor{red}{\ding{55}} & 57.0 & 55.5 & 77.1 & 61.1 \\
      \rowcolor{blue!10} \textbf{+TAI (Ours)} & \textcolor{green!60!black}{\checkmark} & \textbf{59.7} & 58.0 & \textbf{79.6} & \textbf{63.7} \\
      \midrule
      Qwen3-VL-8B & - & \underline{65.5} & \underline{60.6} & \underline{78.8} & \underline{67.4} \\
      +TCD & \textcolor{green!60!black}{\checkmark} & 65.3 & \textbf{61.0} & \textbf{79.1} & \underline{67.4} \\
      +DINO-HEAL & \textcolor{green!60!black}{\checkmark} & 65.4 & 60.2 & \underline{78.8} & 67.3 \\
      \rowcolor{blue!10} \textbf{+TAI (Ours)} & \textcolor{green!60!black}{\checkmark} & \textbf{65.9} & 60.5 & \underline{78.8} & \textbf{67.6} \\
      \midrule
      InternVL2.5-8B & - & 67.8 & \underline{63.5} & \underline{89.0} & \underline{71.6} \\
      +TCD & \textcolor{green!60!black}{\checkmark} & 67.7 & \textbf{64.0} & \underline{89.0} & \underline{71.6} \\
      +DINO-HEAL & \textcolor{green!60!black}{\checkmark} & \textbf{68.3} & 63.0 & \textbf{90.9} & \textbf{72.1} \\
      \rowcolor{blue!10} \textbf{+TAI (Ours)} & \textcolor{green!60!black}{\checkmark} & \underline{68.0} & 63.4 & \underline{89.0} & \underline{71.6} \\
      \bottomrule
    \end{tabular}%
    }
  \end{minipage}
\end{table*}

\subsection{Video Reasoning Benchmarks}
\label{sec:exp_temporal}

\noindent\textbf{Temporal reasoning.} Tab.~\ref{tab:temporal-results} and Tab.~\ref{tab:aotbench-results} report results on three temporal reasoning benchmarks. TAI consistently improves over the baseline across all three models and all three benchmarks without any training, reflecting the shared pattern in the $\hat{\tau}_l$ profile across architectures identified in Sec.~\ref{sec:analysis}. Among training-free methods, TAI achieves the highest average and reaches competitive performance with ArrowRL on Qwen2.5-VL-7B, which requires reinforcement learning with curated temporal rewards and dedicated training. Improvements concentrate on categories sensitive to frame order, such as detecting directional motion or ordering sequential events, while categories that can be resolved from a single frame or remain invariant under reversal receive minimal effect, resulting in consistent overall gains across all benchmarks and models.

\noindent\textbf{General video understanding.} Tab.~\ref{tab:mvbench-results} evaluates whether TAI degrades general understanding capabilities. We use MVBench grouped by relevance to temporal information of the input frames, with detailed grouping and per-subtask results in Appendix~\ref{sec:detailed-results}. TAI improves the temporal-relevant group across all three models, while the temporal-irrelevant and hybrid groups present only minor fluctuations with no decline in overall performance, demonstrating that TAI preserves general video understanding capabilities while improving on temporal reasoning.

\begin{figure}[htbp]
  \centering
  \includegraphics[width=0.9\linewidth]{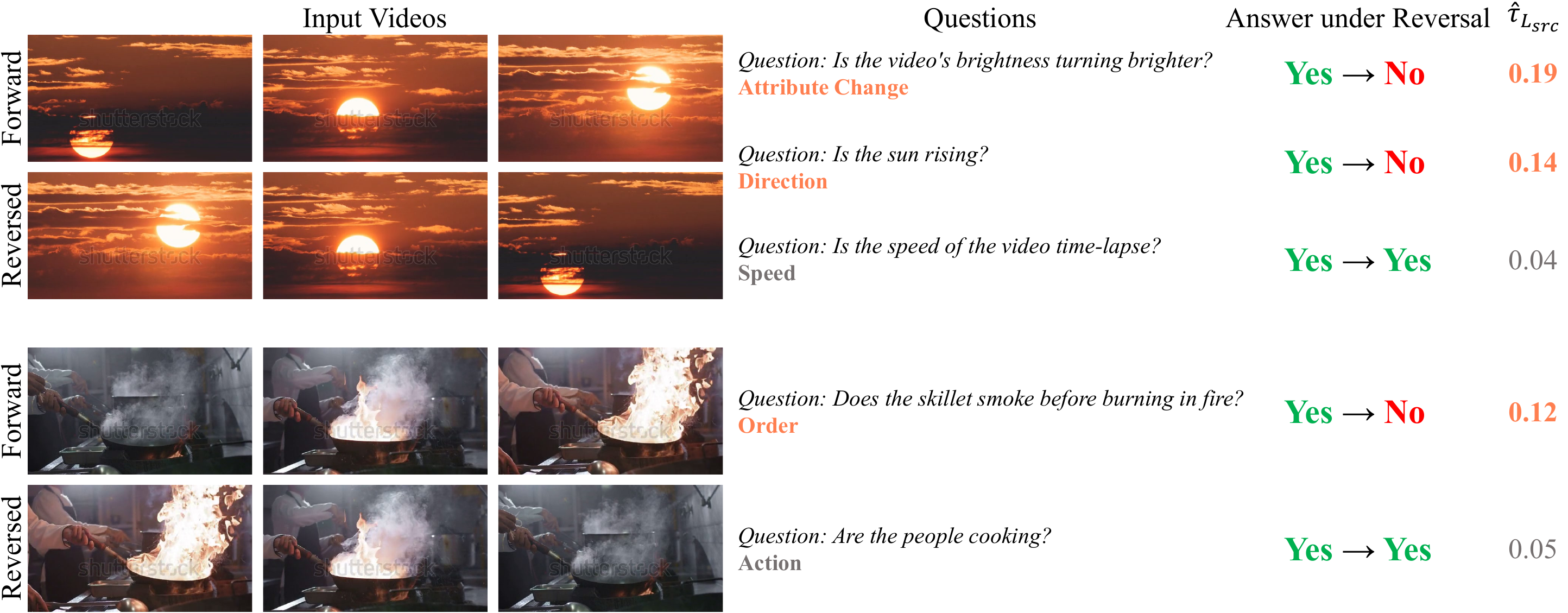}
  \caption{\textbf{TempCompass examples per category on Qwen2.5-VL-7B.} Attribute Change, Direction, and Order produce high $\hat{\tau}_{L_{\mathrm{src}}}$ as their answers change under frame reversal, while Action and Speed produce low $\hat{\tau}_{L_{\mathrm{src}}}$ as they remain invariant. $\hat{\tau}_{L_{\mathrm{src}}}$ scales with reversal sensitivity without supervision.}
  \label{fig:subtask}
\end{figure}

\begin{table}[htbp]
  \caption{\textbf{Steering specificity on TempCompass 
  with Qwen2.5-VL-7B.} Anti-TAI reverses injection 
  direction and selectively degrades reversal-sensitive 
  categories while leaving invariant ones largely 
  unchanged. \colorbox{red!10}{Red} marks 
  reversal-sensitive and \colorbox{orange!10}{orange} 
  moderately sensitive categories.
  ${}^{\dagger}$Per-category mean of $\hat{\tau}_{L_{\mathrm{src}}}$ 
  computed from the temporal divergence profile in 
  Sec.~\ref{sec:analysis}.}
  \label{tab:steering-test}
  \centering
  \small
  \begin{tabular}{l ccccc}
    \toprule
    \textbf{Method} & \textbf{Action} & \textbf{Attr Change} & \textbf{Direction} & \textbf{Order} & \textbf{Speed} \\
    \midrule
    Baseline
      & 94.9 & 77.2 & 59.1 & 75.8 & 59.2 \\
    TAI ($+\beta$)
      & \textcolor{gray}{$-$1.7}
      & \cellcolor{red!10}\textcolor{green!60!black}{$+$6.6}
      & \cellcolor{orange!10}\textcolor{green!60!black}{$+$2.3}
      & \cellcolor{red!10}\textcolor{green!60!black}{$+$3.9}
      & \textcolor{gray}{$+$0.9} \\
    Anti-TAI ($-\beta$)
      & \textcolor{gray}{$-$0.4}
      & \cellcolor{red!10}\textcolor{red}{$-$36.3}
      & \cellcolor{orange!10}\textcolor{red}{$-$12.6}
      & \cellcolor{red!10}\textcolor{red}{$-$27.4}
      & \textcolor{gray}{$-$0.7} \\
    \midrule
    $\hat{\tau}_{L_{\mathrm{src}}}^{\dagger}$
      & 0.059 & \cellcolor{red!10}0.169 & \cellcolor{orange!10}0.074 & \cellcolor{red!10}0.138 & 0.057 \\
    \bottomrule
  \end{tabular}
\end{table}

\subsection{Steering Specificity}
\label{sec:steering}

Fig.~\ref{fig:subtask} illustrates the relationship between frame reversal sensitivity and $\hat{\tau}_{L_{\mathrm{src}}}$ on individual TempCompass examples. Categories whose answers change under reversal, such as Attribute Change, Direction, and Order, produce larger $\hat{\tau}_{L_{\mathrm{src}}}$ values than reversal-invariant categories like Action and Speed. Without any supervision, $\hat{\tau}_{L_{\mathrm{src}}}$ separates time-variant from time-invariant inputs and scales with each input, allowing TAI to inject $\tau_{\mathrm{steer}}$ at a strength matched to the input itself.

Tab.~\ref{tab:steering-test} validates that 
$\tau_{\mathrm{steer}}$ encodes a temporal signal. 
Inverting the sign of $\beta$ reverses the injection 
direction and selectively degrades the three 
reversal-sensitive categories, whose answers change under 
frame reversal, with Attribute Change dropping 36.3\%, 
Order 27.4\%, and Direction 12.6\%, while the two 
reversal-invariant categories, Action and Speed, remain 
within 1\% of baseline under Anti-TAI. The per-category 
$\hat{\tau}_{L_{\mathrm{src}}}$ in the bottom row 
confirms that both the magnitude of improvement and 
degradation track the strength of the extracted temporal 
signal, with the reversal-sensitive categories producing 
larger $\hat{\tau}_{L_{\mathrm{src}}}$ and larger 
performance shifts in both directions than the 
reversal-invariant ones. Among the reversal-sensitive 
categories, Direction produces a relatively smaller 
$\hat{\tau}_{L_{\mathrm{src}}}$ due to heterogeneity 
within the category, as detailed in 
Appendix~\ref{app:within-category}. These results 
establish that $\hat{\tau}_{L_{\mathrm{src}}}$ identifies 
a component of the representation critical for temporal 
order reasoning and that TAI enhances this component, 
enabling the model to better comprehend temporal 
progression in videos.

\subsection{Ablation Studies}
\label{sec:ablation}

\begin{wraptable}{r}{0.30\textwidth}
    \vspace{-17pt}
    \centering
    \small
    \caption{\textbf{Injection schedule ablation} on TempCompass.}
    \label{tab:injection-schedule}
    \setlength{\tabcolsep}{4pt}
    \begin{tabular}{@{}lc@{}}
    \toprule
    \textbf{Schedule} & \textbf{Accuracy} \\
    \midrule
    Baseline              & 73.4 \\
    \midrule
    Uniform ($w_l=1$)     & 75.4 \\
    Reversed profile      & 75.1 \\
    \rowcolor{blue!10} \textbf{Ours} & \textbf{75.6} \\
    \bottomrule
    \end{tabular}
    \vspace{-10pt}
\end{wraptable}

\textbf{Injection schedule.}
Tab.~\ref{tab:injection-schedule} compares schedules for distributing $\tau_{\mathrm{steer}}$ across layers on Qwen2.5-VL-7B. All schedules improve over the baseline, yet their gains differ. A uniform schedule treats every layer equally and does not reflect the natural decay of temporal signal in later layers. Because $\tau_{\mathrm{steer}}$ originates from the representation space of the peak layer, injecting it more strongly into deeper layers where the temporal signal is weaker, as the reversed schedule does, is less aligned with the downstream representations and yields lower gains. Our profile-guided schedule derives $w_l$ directly from each model's $\hat{\tau}_l$ profile, naturally matching the layer-wise signal strength without additional schedule tuning.

\begin{figure}[htbp]
  \centering
  \includegraphics[width=0.99\linewidth]{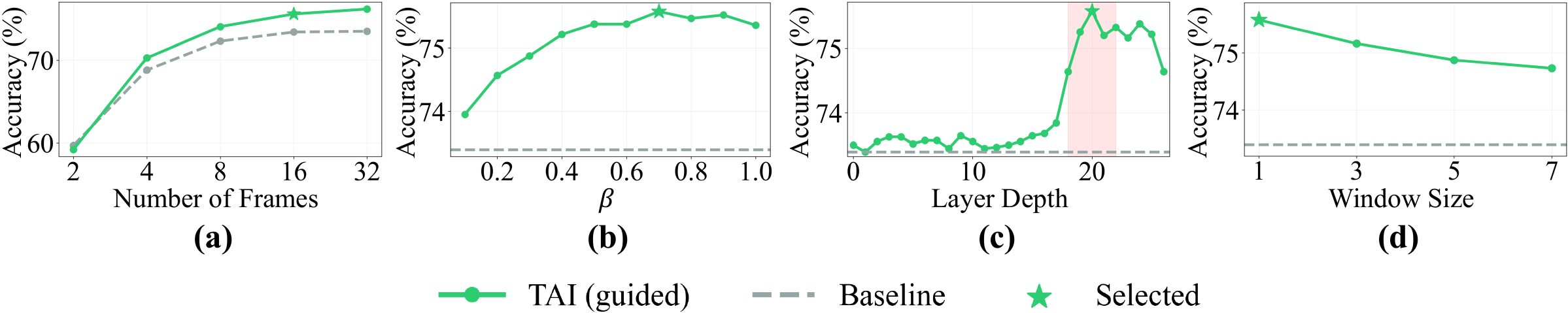}
  \caption{\textbf{Configuration sensitivity of Qwen2.5-VL-7B on TempCompass.} (a) Frame count, (b) injection strength $\beta$, (c) source layer $L_{\mathrm{src}}$, (d) extraction window size $k$.}
  \label{fig:ablation}
  \vspace{-10pt}
\end{figure}

\noindent\textbf{Configuration sensitivity.}
Fig.~\ref{fig:ablation} reports sensitivity to four
configuration choices in TAI. (a)~TAI improves over the 
baseline at every frame count, with only marginal 
difference at 2 frames where the two frames provide 
limited temporal cues for $\tau_{\mathrm{steer}}$, with 
gains growing as more frames provide richer temporal cues 
and saturating beyond 16 frames where additional frames 
yield only marginally richer cues for the short 
TempCompass videos. (b)~Sweeping $\beta$ from $0.1$ to 
$1.0$ in increments of $0.1$, TAI outperforms the 
baseline at every value, with performance increasing 
steadily up to $\beta{=}0.5$ and plateauing beyond. 
(c)~Extracting $\tau_{\mathrm{steer}}$ from the 
$\hat{\tau}_l$ peak layer yields the largest gain, with 
earlier layers producing near-zero improvement and later 
layers showing reduced effect, validating the 
$\hat{\tau}_l$ profile as a reliable indicator of where 
to extract. Together with the profile-guided injection 
weights $w_l$, this leaves $\beta$ as the only free 
hyperparameter in TAI. (d)~Single-layer extraction ($k{=}1$) performs best, yet 
widening the window to $k{=}3$, $5$, or $7$ degrades 
performance only gradually, consistent with the gradual 
shape of the $\hat{\tau}_l$ profile around the peak 
rather than a narrow spike.

\begin{wraptable}{r}{0.51\textwidth}
    \vspace{-18pt}
    \centering
    \small
    \caption{\textbf{Orthogonality of TAI} with other methods.}
    \label{tab:orthogonality}
    \setlength{\tabcolsep}{4pt}
    \begin{tabular}{@{}lccc@{}}
    \toprule
    \textbf{Method} & \textbf{AoTBench} & \textbf{TempCompass} & \textbf{TVBench} \\
    \midrule
    Baseline        & 54.5 & 73.4 & 44.6 \\
    \rowcolor{blue!10}
    \quad + TAI     & \textcolor{green!60!black}{+3.2} & \textcolor{green!60!black}{+2.2} & \textcolor{green!60!black}{+2.0} \\
    \midrule
    ArrowRL         & 58.2 & 75.0 & 46.9 \\
    \rowcolor{blue!10}
    \quad + TAI     & \textcolor{green!60!black}{+2.1} & \textcolor{green!60!black}{+1.5} & \textcolor{green!60!black}{+0.9} \\
    \bottomrule
    \end{tabular}
    \vspace{-10pt}
\end{wraptable}

\textbf{Orthogonality with training-based methods.}
TAI operates at inference time without modifying model
parameters and can therefore be applied on top of any
weight-modified backbone. We verify this by applying TAI
to both the vanilla and ArrowRL-trained
Qwen2.5-VL-7B. As reported in
Tab.~\ref{tab:orthogonality}, TAI yields consistent gains
in both settings. Even after ArrowRL strengthens temporal
reasoning through weight-level optimization, TAI provides
further improvement through activation-level steering,
confirming the two operate through complementary
mechanisms.

\section{Conclusion}
\label{sec:conclusion}

We introduce the temporal divergence vector $\tau_l$ and 
its layer-wise profile, establishing that VideoLLMs 
acquire temporal information at intermediate layers but 
fail to maintain it to the output. The temporal 
divergence profile identifies both where this information 
concentrates and how it decays, providing a natural 
intervention point grounded in the model. Leveraging this 
profile, we propose Temporal Activation Injection (TAI), 
a training-free method that extracts 
$\tau_{\mathrm{steer}}$ at the peak and reinjects it into 
subsequent layers following the decay reflected in the 
profile, compensating for the signal that the model 
produces but does not preserve. TAI adapts to each input 
through the magnitude of $\tau_{\mathrm{steer}}$, 
selectively improving tasks dependent on temporal 
information while leaving non-temporal capabilities 
largely unaffected. Experiments across three VideoLLMs 
and four benchmarks confirm this. The $\hat{\tau}_l$ 
profile determines both the extraction layer and the 
injection schedule from the model, leaving 
$\beta$ as the only free hyperparameter. TAI provides 
additional gains on top of training-based methods, 
confirming orthogonality to weight-level optimization. In 
VideoLLMs, temporal information concentrates at 
intermediate layers yet diminishes before reaching the 
output, and leveraging it before it fades proves 
sufficient to improve the temporal reasoning essential 
for video understanding.

\section{Limitations}
\label{sec:limitations}

This work focuses on temporal information arising from 
the sequential ordering of frames. While this covers 
important aspects such as event ordering, directional 
motion, and attribute change over time, video 
understanding involves additional factors including 
spatial reasoning, object interactions, and audio-visual 
correspondence that fall outside the scope of this work. 
The $\hat{\tau}_l$ framework could in principle be 
extended to capture other axes of variation by designing 
appropriate contrastive pairs beyond frame order reversal, 
but exploring such extensions remains future work.

\newpage

\bibliographystyle{plainnat}
\bibliography{references}    

\newpage
\appendix
\section*{Appendix}
\setcounter{figure}{0}
\setcounter{table}{0}
\renewcommand{\thefigure}{\Alph{figure}}
\renewcommand{\thetable}{\Alph{table}}

\newcommand{\apptocgroup}[1]{\vspace{6pt}\noindent\textbf{#1}\par\vspace{2pt}}
\newcommand{\apptocentry}[1]{\noindent\hspace{1em}\hyperref[#1]{\makebox[1.6em][l]{\ref*{#1}}\nameref*{#1}}\leaders\hbox to 0.6em{\hss.\hss}\hfill\pageref*{#1}\par}
\apptocgroup{Analysis and Profile Estimation}
\apptocentry{sec:suppl_why_layer}
\apptocentry{app:npairs}
\apptocgroup{Additional Results}
\apptocentry{app:more-models}
\apptocentry{app:robustness}
\apptocentry{sec:detailed-results}
\apptocentry{app:generation}
\apptocentry{app:within-category}
\apptocentry{app:aotbench-adaptive}
\apptocgroup{Implementation Details}
\apptocentry{app:config}
\apptocentry{app:frame-count}
\apptocentry{sec:reproduction}
\apptocentry{app:implementation}
\apptocgroup{Visualization Details and Examples}
\apptocentry{app:fig1-details}
\apptocentry{sec:per-benchmark-examples}
\clearpage

\section{Justification for Layer-Level Last-Token Extraction}
\label{sec:suppl_why_layer}

\begin{figure*}[htbp]
    \centering
    \includegraphics[width=\textwidth]{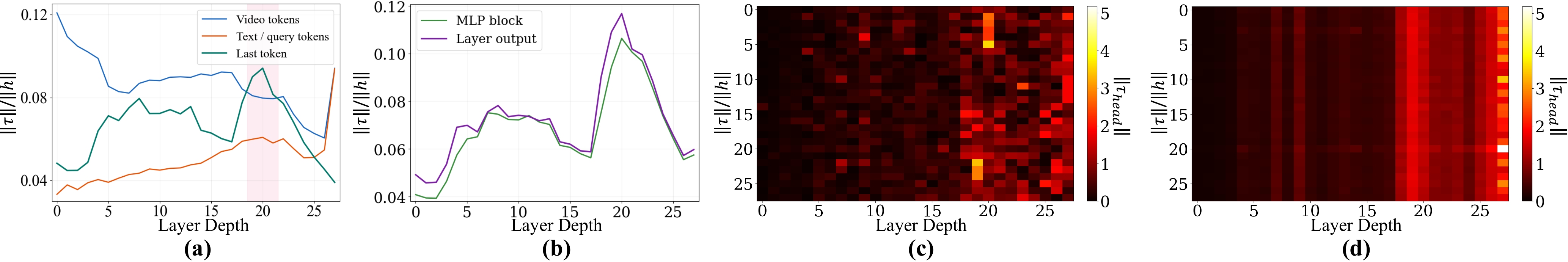}
    \caption{\textbf{Temporal divergence decomposition across token positions and sub-layer computations.} \textbf{(a)}~Token-group $\hat{\tau}_l$ profiles. \textbf{(b)}~$\hat{\tau}_l$ comparison between MLP block and layer output. \textbf{(c,\,d)}~Per-head temporal sensitivity before \textbf{(c)} and after \textbf{(d)} the attention output projection.}
    \label{fig:suppl_why_layer}
\end{figure*}

The analysis in Sec.~\ref{sec:analysis} measures temporal divergence $\hat{\tau}_l$ exclusively at the last-token hidden state of each layer output. Fig.~\ref{fig:suppl_why_layer} presents additional experiments on Qwen2.5-VL-7B~\citep{bai2025qwen25vltechnicalreport} that validate this design choice across three dimensions: token position (Sec.~\ref{sec:suppl_token_pos}), sub-layer computation (Sec.~\ref{sec:suppl_sublayer}), and attention head granularity (Sec.~\ref{sec:suppl_head}).

\subsection{Temporal Information Transfers from Video to Text Tokens}
\label{sec:suppl_token_pos}

We extend the $\hat{\tau}_l$ measurement to video tokens, text tokens, and the last token separately, as illustrated in Fig.~\ref{fig:suppl_why_layer} (a). Video tokens exhibit high divergence from the earliest layers, as temporally reversed videos contain different visual content. Text tokens receive identical input across the forward and reversed conditions, yet their divergence increases steadily across layers. This progressive increase directly demonstrates that the attention mechanism transfers temporal information from video tokens into text representations. Under causal attention, the last token is the only position that attends to all preceding video and text tokens, making it a natural aggregation point for the temporal signal transferred from video to text representations. Accordingly, its $\hat{\tau}_l$ rises as both video and text representations accumulate temporal information, peaking at Layer~20 as measured in Sec.~\ref{sec:tau_def}. The peak observed in the main analysis therefore reflects the culmination of this layer-wise temporal processing pipeline, consistent with recent analyses of cross-modal information flow in VideoLLMs~\citep{kim2026map, shi2026causality}.

\subsection{Layer Output Retains the Most Complete Temporal Signal}
\label{sec:suppl_sublayer}
We compare the temporal divergence of the MLP block and the full layer output, as illustrated in Fig.~\ref{fig:suppl_why_layer} (b). Both exhibit similar profile shapes, yet the full layer output consistently maintains higher magnitude. This gap indicates that the full layer output benefits from the entire computation pipeline, where the attention block extracts temporal information and the MLP further processes it, rather than either component alone. We therefore extract the steering vector from the full layer output, as it retains the most complete temporal signal.

\subsection{Head-Level Temporal Sensitivity Dissolves After Projection}
\label{sec:suppl_head}
We examine per-head temporal sensitivity before and after the attention output projection $W_O$, as illustrated in Fig.~\ref{fig:suppl_why_layer} (c) and (d). We measure the per-head temporal divergence magnitude, $\|\tau_{\mathrm{head}}\|$, for each attention head independently. Before $W_O$, a small number of heads at the peak layers exhibit notably higher $\|\tau_{\mathrm{head}}\|$, indicating that certain heads respond more strongly to temporal differences in the input. After $W_O$ projects the concatenated head outputs into the hidden representation passed to subsequent layers, this concentration largely disappears and the signal becomes uniformly distributed across heads. All subsequent computation operates on this post-projection representation, meaning that even if representation engineering were applied at the head level, the injected signal would be flattened by the learned projection before reaching downstream layers. Layer-level extraction avoids this bottleneck and retains the full temporal signal as composed by the model.

\section{Robustness of $\hat{\tau}_l$ Profile Estimation}
\label{app:npairs}

\begin{figure}[t]
  \centering
  \includegraphics[width=0.87\linewidth]{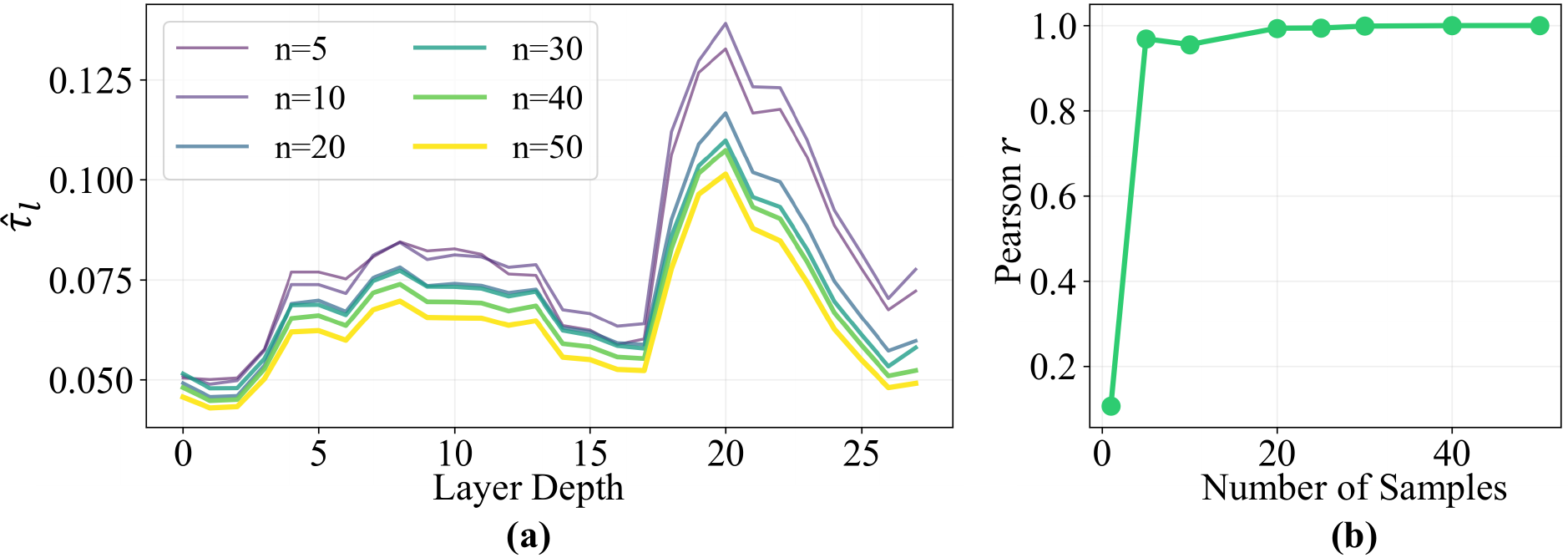}
  \caption{\textbf{$\hat{\tau}_l$ profile stability on Qwen2.5-VL-7B as the number of video pairs varies.} \textbf{(a)}~Overlaid profiles at varying $n$. \textbf{(b)}~Pearson correlation with the $n{=}50$ reference profile.}
  \label{fig:npairs}
\end{figure}

The temporal divergence profile used throughout this work 
is computed from 50 forward-reverse video pairs with 
binary temporal questions from TempCompass. 
Fig.~\ref{fig:npairs} examines how the profile changes as 
the number of pairs varies. The overlay in 
Fig.~\ref{fig:npairs}(a) shows that the characteristic 
peak-then-decay shape emerges with as few as 5 pairs, and 
the Pearson correlation with the 50-pair profile in 
Fig.~\ref{fig:npairs}(b) exceeds 0.95 at $n{=}5$ and 
reaches near-perfect agreement by $n{=}20$. The profile 
shape is therefore robust to the number of pairs, 
confirming that it captures a structural property of the 
model rather than properties of specific videos. We use 
50 pairs to ensure stable estimates for downstream 
analyses that depend on precise layer-wise values, such 
as the attention knockout experiment in 
Sec.~\ref{sec:knockout}.

\begin{table}[htbp]
  \centering
  \caption{Cross-domain $\hat{\tau}_l$ profile estimation on Qwen2.5-VL-7B.}
  \label{tab:cross-domain}
  \small
  \begin{tabular}{@{}lccccc@{}}
    \toprule
    \textbf{Profile Source} & \textbf{Runs} & \textbf{Peak} & \textbf{TempCompass} & \textbf{TVBench} & \textbf{AoTBench} \\
    \midrule
    TempCompass & 1 & L20 & 75.58 & 46.58 & 57.70 \\
    TVBench     & 9 & L20 & -0.15 & 0.00 & 0.00 \\
    AoTBench    & 9 & L20 & -0.11 & -0.19 & +0.28 \\
    \bottomrule
  \end{tabular}
\end{table}

Beyond the number of pairs, we further examine whether the profile depends on the source of the videos. On Qwen2.5-VL-7B, we re-estimate the $\hat{\tau}_l$ profile from TVBench and AoTBench videos with $n \in \{10, 30, 50\}$ pairs and three random seeds each. Since TVBench contains only multi-choice questions, we use a standardized yes/no temporal probe ``Is this video playing in reverse?'' across all domains, consistent with the binary temporal questions used in Sec.~\ref{sec:tau_def}. As shown in Tab.~\ref{tab:cross-domain}, the peak consistently falls at L20 in all 18 additional runs. To test whether the source affects TAI itself, we replace the TempCompass profile with the TVBench or AoTBench profile estimated from 50 pairs with the first seed, which leaves accuracy nearly unchanged on every benchmark. Tab.~\ref{tab:cross-domain} reports these differences relative to the TempCompass profile to two decimal places, since rounding to one decimal place would distort such small differences. The profile therefore captures a structural property of the model rather than of the source videos, and requires no per-domain recalibration.

\section{Generalization to Additional Model Families}
\label{app:more-models}

\begin{figure}[htbp]
  \centering
  \includegraphics[width=0.87\linewidth]{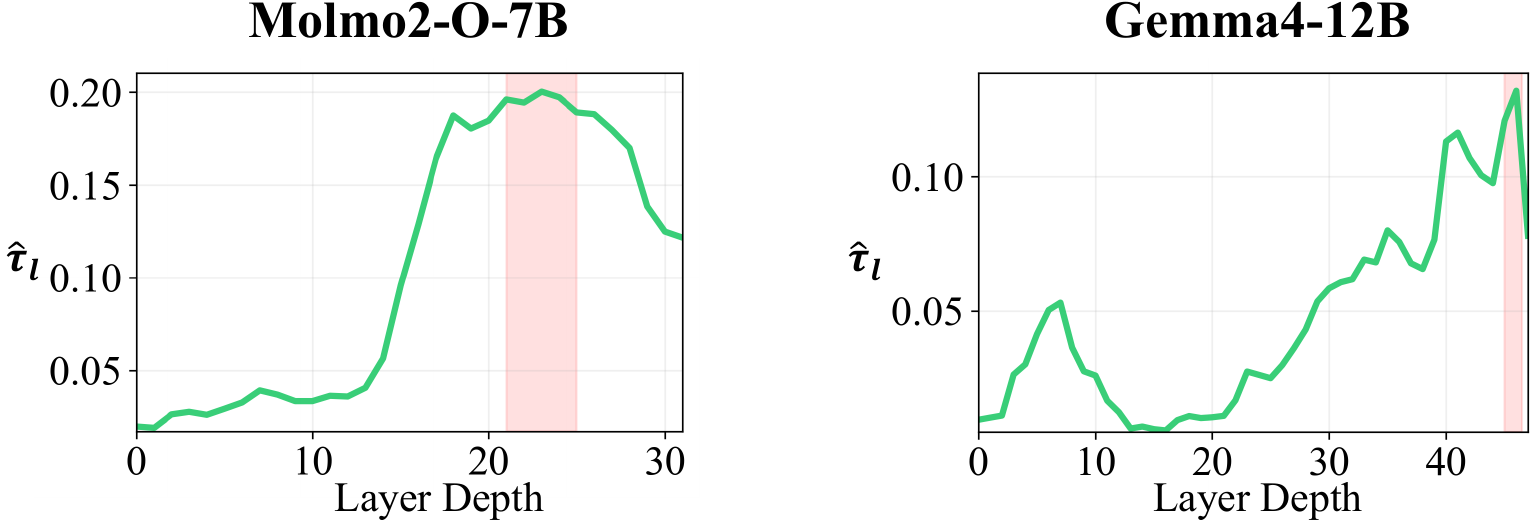}
  \caption{\textbf{$\hat{\tau}_l$ profiles on additional model families.} Red shaded regions mark the peak layers.}
  \label{fig:more-models-profile}
\end{figure}

We further examine TAI on Molmo2-O-7B and Gemma4-12B, two model families not addressed in the main paper. Molmo2-O-7B is built on the OLMo backbone with 32 layers, and Gemma4-12B is a natively multimodal model with 48 layers. For both models, we extract the $\hat{\tau}_l$ profile following Sec.~\ref{sec:tau_def} and evaluate TAI on TempCompass with 16-frame input and $\beta{=}0.3$. As shown in Fig.~\ref{fig:more-models-profile} and Tab.~\ref{tab:more-models}, both models exhibit the peak-then-decay profile, and TAI improves over the baseline on both. Even on these recent models, simply locating the $\hat{\tau}_l$ peak as $L_{\mathrm{src}}$, extracting $\tau_{\mathrm{steer}}$ there, and injecting it into the subsequent layers yields performance gains without further modification to these models.

\begin{table}[htbp]
  \centering
  \caption{TempCompass results of TAI on Molmo2-O-7B and Gemma4-12B, with $L_{\mathrm{src}}$ set at the $\hat{\tau}_l$ peak of each model. TAI improves over the baseline on both models.}
  \label{tab:more-models}
  \small
  \begin{tabular}{@{}llccccc@{}}
    \toprule
    \textbf{Model} & \textbf{LM Backbone} & \textbf{\#Layers} & \textbf{$L_{\mathrm{src}}$} & \textbf{Baseline} & \textbf{+TAI} & \textbf{$\Delta$} \\
    \midrule
    Molmo2-O-7B & OLMo   & 32 & 23 & 75.8 & 77.6 & +1.8 \\
    Gemma4-12B  & Native & 48 & 46 & 70.6 & 72.4 & +1.8 \\
    \bottomrule
  \end{tabular}
\end{table}

\section{Robustness of TAI Gains}
\label{app:robustness}

Each reported accuracy is deterministic for a fixed input since all methods use greedy decoding. We therefore examine whether the gains of TAI hold under input perturbations on TempCompass with Qwen2.5-VL-7B, applying the paired McNemar test to each comparison. For frame sampling, we perturb each sampled frame position around the uniform grid with different random seeds and evaluate all TempCompass formats. For question phrasing, we evaluate the yes/no format with two prompt variants, Prompt A \textit{Based on the video, \{question\}} and Prompt B \textit{Could you tell me: \{question\}}, since the other formats embed answer options within the question. As shown in Tab.~\ref{tab:robust-sampling} and Tab.~\ref{tab:robust-prompt}, TAI improves over the baseline under every perturbation with statistical significance. TAI also improves at every number of sampled frames, as presented in Fig.~\ref{fig:ablation}(a).

\newsavebox{\robustSamplingBox}
\newsavebox{\robustPromptBox}
\sbox{\robustSamplingBox}{\small\setlength{\tabcolsep}{3pt}%
    \begin{tabular}{lcccc}
      \toprule
      \textbf{Sampling} & \textbf{Baseline} & \textbf{TAI} & \textbf{$\Delta$} & \textbf{McNemar $p$} \\
      \midrule
      Uniform & 73.39 & 75.58 & +2.19 & $<10^{-6}$ \\
      Seed 0  & 73.50 & 75.61 & +2.11 & $<10^{-6}$ \\
      Seed 1  & 73.52 & 76.16 & +2.64 & $<10^{-6}$ \\
      \midrule
      Mean    & 73.47$_{\pm 0.06}$ & 75.78$_{\pm 0.27}$ & +2.31$_{\pm 0.23}$ & -- \\
      \bottomrule
    \end{tabular}}
\sbox{\robustPromptBox}{\small\setlength{\tabcolsep}{3pt}%
    \begin{tabular}{lcccc}
      \toprule
      \textbf{Prompt} & \textbf{Baseline} & \textbf{TAI} & \textbf{$\Delta$} & \textbf{McNemar $p$} \\
      \midrule
      Original & 72.5 & 75.5 & +3.0 & $<10^{-4}$ \\
      Prompt A & 72.7 & 76.4 & +3.7 & $<10^{-6}$ \\
      Prompt B & 73.4 & 75.6 & +2.2 & $0.004$ \\
      \midrule
      Mean     & 72.9$_{\pm 0.4}$ & 75.8$_{\pm 0.4}$ & +3.0$_{\pm 0.6}$ & -- \\
      \bottomrule
    \end{tabular}}

\begin{table}[htbp]
  \centering
  \begin{minipage}[t]{0.49\linewidth}
    \centering
    \caption{Robustness of TAI to frame sampling.}
    \label{tab:robust-sampling}
    \resizebox{\linewidth}{!}{\usebox{\robustSamplingBox}}
  \end{minipage}\hfill
  \begin{minipage}[t]{0.49\linewidth}
    \centering
    \caption{Robustness of TAI to question phrasing.}
    \label{tab:robust-prompt}
    \resizebox{\dimexpr\linewidth*\wd\robustPromptBox/\wd\robustSamplingBox\relax}{!}{\usebox{\robustPromptBox}}
  \end{minipage}
\end{table}

\section{Detailed Empirical Results}
\label{sec:detailed-results}

We group MVBench subtasks by whether their ground-truth answers depend on temporal information of the input frames, following the criterion described in Sec.~\ref{sec:per-benchmark-examples}.

\begin{itemize}[leftmargin=*, itemsep=2pt, parsep=0pt]
\item \textbf{Temporal-relevant} (10 tasks): Action Antonym, Action Localization, Action Prediction, Action Sequence, Character Order, Egocentric Navigation, Moving Direction, Object Shuffle, Scene Transition, State Change.
\item \textbf{Temporal-irrelevant} (4 tasks): Action Count, Fine-grained Action, Object Interaction, Unexpected Action.
\item \textbf{Hybrid} (4 tasks): Counterfactual Inference, Moving Attribute, Moving Count, Object Existence. These tasks originate from the CLEVRER dataset and contain a mixture of temporal-dependent and temporal-independent questions within each subtask.
\end{itemize}

Please refer to Tab.~\ref{tab:mvbench-results-full} for the full per-subtask breakdown.

\begin{table*}[htbp]
  \caption{General video understanding results on MVBench subtasks. \textbf{Bold} marks the best and \underline{underline} the second-best per group; \colorbox{blue!10}{blue} highlights our method. All methods use 16-frame input.}
  \label{tab:mvbench-results-full}
  \centering
  \resizebox{\textwidth}{!}{%
  \begin{tabular}{lc cccccccccccccccccc c}
    \toprule
    \textbf{Models} & \textbf{\shortstack{Training\\-Free}}
    & \rotatebox{70}{Act. Antonym}
    & \rotatebox{70}{Act. Count}
    & \rotatebox{70}{Act. Local.}
    & \rotatebox{70}{Act. Pred.}
    & \rotatebox{70}{Act. Seq.}
    & \rotatebox{70}{Char. Order}
    & \rotatebox{70}{Counterfact.}
    & \rotatebox{70}{Ego. Nav.}
    & \rotatebox{70}{Fine Act.}
    & \rotatebox{70}{Mov. Attr.}
    & \rotatebox{70}{Mov. Count}
    & \rotatebox{70}{Mov. Dir.}
    & \rotatebox{70}{Obj. Exist.}
    & \rotatebox{70}{Obj. Inter.}
    & \rotatebox{70}{Obj. Shuf.}
    & \rotatebox{70}{Scene Trans.}
    & \rotatebox{70}{State Chg.}
    & \rotatebox{70}{Unexp. Act.}
    & \textbf{AVG} \\
    \midrule
    Qwen2.5-VL-7B & - & 74.5 & \underline{46.5} & 41.0 & 58.0 & 67.0 & \underline{73.5} & \underline{70.5} & \underline{32.0} & 46.0 & 91.0 & 62.5 & \underline{52.5} & 86.9 & \textbf{67.5} & \textbf{44.0} & \textbf{91.0} & \underline{54.5} & \underline{74.0} & 62.9 \\
    +TCD & \textcolor{green!60!black}{\checkmark} & \underline{76.0} & 46.0 & \underline{42.0} & \textbf{61.0} & \underline{70.2} & \underline{73.5} & 66.5 & 31.0 & 46.0 & \underline{92.0} & 63.5 & 52.0 & \underline{88.9} & \underline{66.0} & \underline{43.0} & 89.0 & \underline{54.5} & \underline{74.0} & 63.0 \\
    +DINO-HEAL & \textcolor{green!60!black}{\checkmark} & 73.5 & \textbf{47.0} & \textbf{43.0} & \underline{60.5} & 67.6 & \textbf{74.0} & \textbf{71.0} & \textbf{33.5} & \textbf{47.0} & \underline{92.0} & 65.0 & 49.5 & 85.4 & 65.5 & 42.5 & \underline{90.0} & 53.5 & \textbf{75.0} & \underline{63.1} \\
    +ArrowRL & \textcolor{red}{\ding{55}} & 74.5 & 42.5 & 41.0 & 60.0 & \underline{70.2} & 68.0 & 64.0 & 31.5 & 45.5 & \underline{92.0} & \textbf{67.0} & 52.0 & 85.4 & \underline{66.0} & 31.5 & 88.0 & 54.0 & 68.0 & 61.1 \\
    \rowcolor{blue!10} +TAI (Ours) & \textcolor{green!60!black}{\checkmark} & \textbf{77.5} & 46.0 & \underline{42.0} & \underline{60.5} & \textbf{70.7} & \underline{73.5} & 70.0 & \textbf{33.5} & \underline{46.5} & \textbf{92.5} & \underline{66.5} & \textbf{53.0} & \textbf{89.4} & 64.5 & 41.0 & \textbf{91.0} & \textbf{55.0} & \textbf{75.0} & \textbf{63.7} \\
    \midrule
    Qwen3-VL-8B & - & \underline{81.5} & \underline{41.0} & 41.0 & 70.5 & \underline{76.1} & 76.5 & \textbf{70.5} & \underline{38.0} & \textbf{49.5} & \underline{92.0} & 68.5 & \underline{64.5} & \underline{84.3} & 71.0 & \textbf{41.5} & \textbf{94.0} & \underline{72.5} & \textbf{81.0} & \underline{67.4} \\
    +TCD & \textcolor{green!60!black}{\checkmark} & 79.5 & \textbf{43.5} & \underline{41.5} & \textbf{72.0} & 75.5 & \underline{78.0} & \underline{70.0} & \underline{38.0} & 48.5 & \underline{92.0} & \underline{70.0} & 63.5 & \underline{84.3} & \textbf{72.5} & 39.0 & \underline{93.0} & \textbf{73.0} & 79.5 & \underline{67.4} \\
    +DINO-HEAL & \textcolor{green!60!black}{\checkmark} & \textbf{82.5} & 39.5 & \textbf{42.0} & \underline{71.5} & \textbf{76.6} & 77.5 & \underline{70.0} & 37.0 & \underline{49.0} & 91.5 & 69.0 & 62.5 & \textbf{84.8} & \underline{72.0} & \textbf{41.5} & \underline{93.0} & 71.0 & \underline{80.5} & 67.3 \\
    \rowcolor{blue!10} +TAI (Ours) & \textcolor{green!60!black}{\checkmark} & \textbf{82.5} & \textbf{43.5} & \underline{41.5} & 68.0 & 75.5 & \textbf{81.0} & 66.0 & \textbf{40.5} & 47.5 & \textbf{93.0} & \textbf{72.0} & \textbf{68.0} & \underline{84.3} & 70.0 & \underline{39.5} & 92.0 & 71.0 & \textbf{81.0} & \textbf{67.6} \\
    \midrule
    InternVL2.5-8B & - & \underline{89.0} & \underline{47.0} & \textbf{46.0} & 77.5 & \underline{83.0} & \textbf{77.5} & \underline{79.0} & \underline{40.5} & \textbf{45.0} & \underline{98.5} & \underline{82.0} & \underline{77.0} & 96.5 & \underline{86.0} & \underline{38.5} & \underline{92.0} & 58.0 & \textbf{76.0} & \underline{71.6} \\
    +TCD & \textcolor{green!60!black}{\checkmark} & \underline{89.0} & \textbf{50.5} & 44.5 & 77.5 & \underline{83.0} & 76.0 & 78.5 & \underline{40.5} & \underline{44.5} & \underline{98.5} & \underline{82.0} & 75.5 & 97.0 & 85.0 & \textbf{39.0} & \underline{92.0} & \textbf{60.0} & \textbf{76.0} & \underline{71.6} \\
    +DINO-HEAL & \textcolor{green!60!black}{\checkmark} & \textbf{90.0} & \underline{47.0} & 45.0 & \underline{78.0} & \textbf{84.0} & \textbf{77.5} & \textbf{80.0} & 39.0 & 44.0 & \textbf{99.0} & \textbf{86.5} & \textbf{82.0} & \textbf{98.0} & 85.0 & \underline{38.5} & \underline{92.0} & 58.0 & \textbf{76.0} & \textbf{72.1} \\
    \rowcolor{blue!10} +TAI (Ours) & \textcolor{green!60!black}{\checkmark} & \textbf{90.0} & 46.5 & \underline{45.5} & \textbf{79.0} & 82.4 & \underline{77.0} & \textbf{80.0} & \textbf{42.0} & \underline{44.5} & 97.0 & 81.5 & 74.5 & \underline{97.5} & \textbf{86.5} & \textbf{39.0} & \textbf{92.5} & \underline{58.5} & \textbf{76.0} & \underline{71.6} \\
    \bottomrule
  \end{tabular}%
  }
  \vspace{-10pt}
\end{table*}

\section{Free-Form Generation and Injection Position}
\label{app:generation}
\begin{figure}[htbp]
  \centering
  \includegraphics[width=0.96\linewidth]{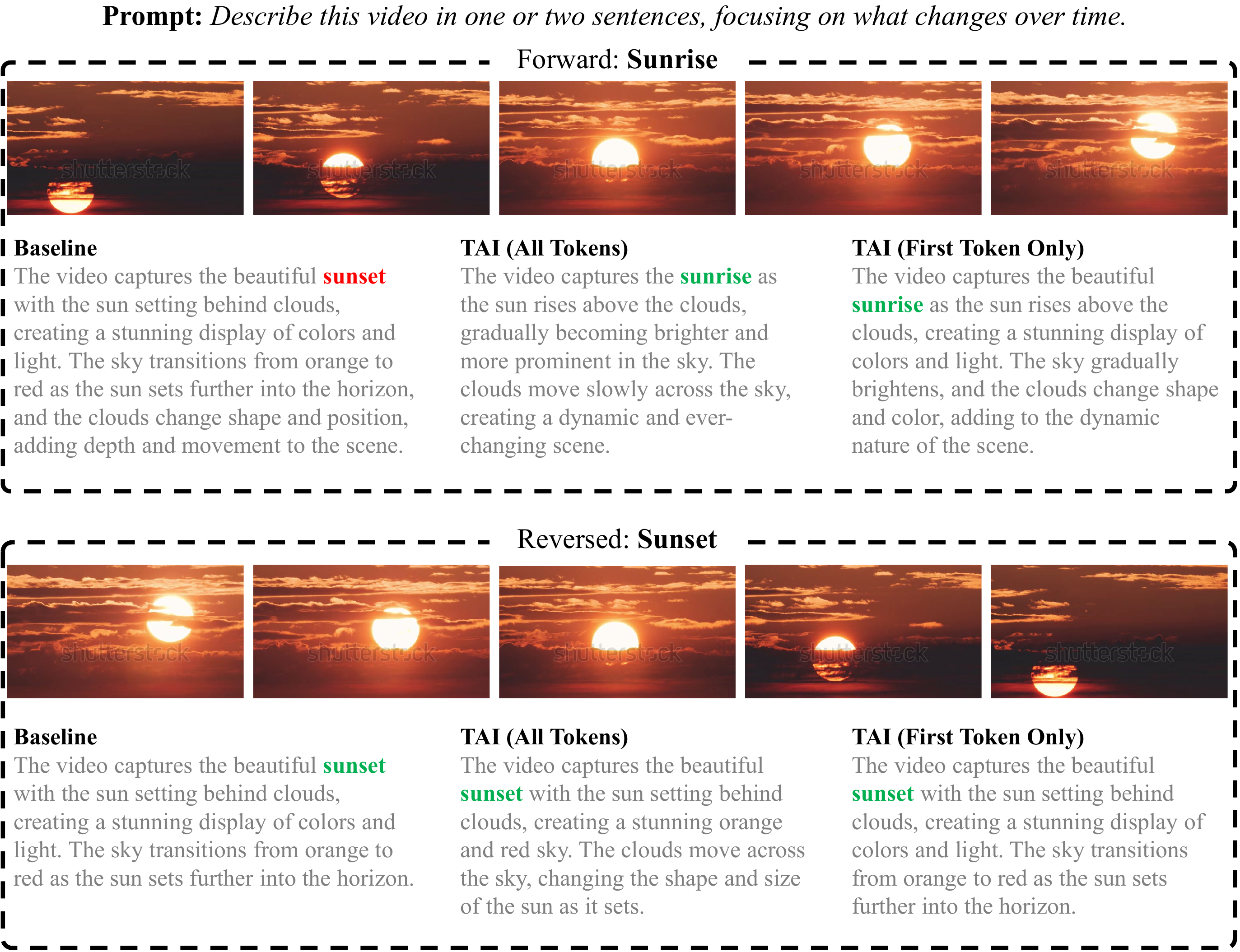}
  \caption{\textbf{Free-form generation under different injection positions.} Responses to a sunrise video and its reversal. Correct and incorrect directions are marked in \textcolor{qualgreen}{\textbf{green}} and \textcolor{qualred}{\textbf{red}}.}
  \label{fig:generation-quality}
\end{figure}

Beyond the single-token answers of the benchmarks, we examine whether TAI carries over to free-form generation. For such answers, TAI injects $\tau_{\mathrm{steer}}$ once at the last token where the answer is produced, whereas free-form generation produces a response token by token, so we qualitatively examine TAI under two injection positions. In the all-token setting, $\tau_{\mathrm{steer}}$ is injected at every decoding step so that each generated token receives the temporal signal directly. In the first-token setting, it is injected only when generating the first token, and the remaining tokens receive the temporal signal only through attention to the steered representation of that token. We generate responses with Qwen2.5-VL-7B to a sunrise video and its reversed counterpart, which depicts a sunset. As shown in Fig.~\ref{fig:generation-quality}, the baseline describes both videos as a sunset, although the forward video depicts a sunrise. TAI describes each direction correctly under both injection positions, capturing the temporal direction regardless of where $\tau_{\mathrm{steer}}$ is injected. The generated responses also remain comparable to the baseline in fluency and detail.

\section{Within-Category Variation of $\hat{\tau}_{L_{\mathrm{src}}}$}
\label{app:within-category}

\begin{figure}[htbp]
\centering
\includegraphics[width=\textwidth]{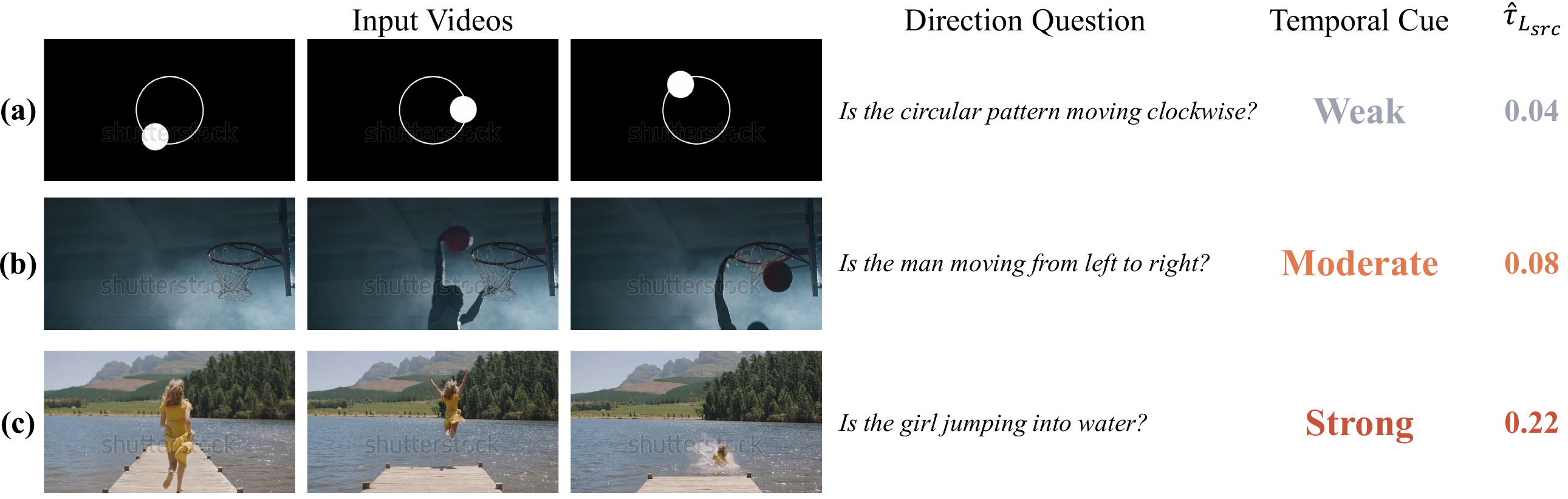}
\caption{\textbf{$\hat{\tau}_{L_{\mathrm{src}}}$ in the Direction category varies with the strength of the temporal cue.} Each question is (a) uncertain, (b) ambiguous, and (c) clear to answer, and $\hat{\tau}_{L_{\mathrm{src}}}$ corresponds to this cue.}
\label{fig:direction-examples}
\end{figure}

\subsection{Heterogeneity in TempCompass Direction Category}
\label{app:direction-decomp}

The smaller average $\hat{\tau}_{L_{\mathrm{src}}}$ on Direction relative to Attribute Change and Order in Tab.~\ref{tab:steering-test} reflects internal heterogeneity within the category. Fig.~\ref{fig:direction-examples} shows three Direction questions whose $\hat{\tau}_{L_{\mathrm{src}}}$ differs by nearly an order of magnitude despite all belonging to the same category. In Fig.~\ref{fig:direction-examples}(a), the circular pattern rotates so quickly that the sampled frames alone are insufficient to determine whether the rotation is clockwise, leaving the temporal cue for reversal weak and yielding $\hat{\tau}_{L_{\mathrm{src}}}=0.04$. In Fig.~\ref{fig:direction-examples}(b), the man is a basketball player jumping to dunk, and the dominant motion is vertical rather than horizontal, leaving the left-to-right direction only weakly present and producing a moderate $\hat{\tau}_{L_{\mathrm{src}}}=0.08$. In Fig.~\ref{fig:direction-examples}(c), the girl is clearly jumping into water, and the reversed video depicts the opposite event of jumping out of water rather than into it, producing $\hat{\tau}_{L_{\mathrm{src}}}=0.22$. The Direction average in Tab.~\ref{tab:steering-test} therefore lands between Attribute Change and Order on the high side and Action and Speed on the low side, because the category mixes weakly cued cases with strongly cued ones. Thus, measuring $\hat{\tau}_{L_{\mathrm{src}}}$ per input and modulating the steering strength accordingly is essential for TAI, as it matches the injection magnitude to the temporal cue actually present in each video and question.

\subsection{Marginal Gains on Temporal-Relevant Tasks}
\label{app:mvbench-decomp}

The improvement from TAI on the temporal-relevant group of MVBench in Tab.~\ref{tab:mvbench-results} is consistently positive but smaller than on TempCompass. The same input-level mechanism analyzed in Sec.~\ref{app:direction-decomp} extends to the task level. While MVBench tasks such as Action Sequence and Moving Direction belong to temporal-relevant categories by topic, the underlying videos and questions vary in how strongly the temporal cue is present. We expect $\hat{\tau}_{L_{\mathrm{src}}}$ on these inputs to be on average smaller than on TempCompass questions where the answer flips under reversal in a controlled manner, leading to smaller injection magnitudes and thus smaller absolute gains.

\section{Adaptive Injection Magnitude on AoTBench}
\label{app:aotbench-adaptive}

AoTBench consists of five subtasks that differ in how 
their input videos respond to temporal reversal. 
Rtime\textsubscript{v2t} and AoTBench\textsubscript{QA} 
present single-direction videos where reversing the frame 
order changes the temporal content, producing a 
meaningful $\tau_{\mathrm{steer}}$ and enabling TAI to 
improve performance. In contrast, ReverseFilm, UCF101, 
and Rtime\textsubscript{t2v} construct each input by 
concatenating a forward video clip $F$ with its reversed 
copy through a black frame, yielding 
$V = [F, \mathbf{0}, \mathrm{rev}(F)]$ as illustrated in Fig.~\ref{fig:aot-examples-2}. This structure is 
invariant under temporal reversal, as 
$\mathrm{rev}(V) = V$, though uniform frame sampling may 
not produce a perfectly palindromic sequence, resulting 
in small but nonzero $\tau_{\mathrm{steer}}$. Tab.~\ref{tab:aot-tau} confirms 
this separation, with reversal-sensitive subtasks 
producing $\hat{\tau}_{L_{\mathrm{src}}}$ an order of 
magnitude larger than reversal-invariant ones. TAI 
therefore applies near-zero injection on the invariant 
subtasks without task-specific conditioning, 
demonstrating that the input-level adaptation in 
Sec.~\ref{sec:tai} naturally distinguishes between 
reversal-sensitive and invariant inputs.

\begin{table}[htbp]
  \centering
  \caption{\textbf{Per-subtask $\hat{\tau}_{L_{\mathrm{src}}}$ 
  on AoTBench} with Qwen2.5-VL-7B. Reversal-invariant 
  subtasks produce near-zero 
  $\hat{\tau}_{L_{\mathrm{src}}}$, and TAI applies 
  correspondingly minimal injection.}
  \label{tab:aot-tau}
  \small
  \begin{tabular}{l ccc | cc}
    \toprule
    \textbf{Subtask} & \textbf{ReverseFilm} & \textbf{UCF101} & \textbf{Rtime\textsubscript{t2v}} & \textbf{Rtime\textsubscript{v2t}} & \textbf{AoTBench\textsubscript{QA}} \\
    \midrule
    Reversal & \multicolumn{3}{c|}{Invariant} & \multicolumn{2}{c}{Sensitive} \\
    \midrule
    $\hat{\tau}_{L_{\mathrm{src}}}$ & 0.042 & 0.024 & 0.033 & \cellcolor{red!10}\textbf{0.190} & \cellcolor{red!10}\textbf{0.172} \\
    \bottomrule
  \end{tabular}
\end{table}

\section{Analysis and Experiment Configuration}
\label{app:config}
All analyses and experiments are conducted on a single 
NVIDIA A6000 GPU. The attention knockout experiments in 
Sec.~\ref{sec:knockout} use a sliding window of $k_{\mathrm{ak}}$ 
layers for stability, with $k_{\mathrm{ak}}{=}3$ for Qwen2.5-VL-7B 
(28 layers), $k_{\mathrm{ak}}{=}9$ for Qwen3-VL-8B (36 layers), and 
$k_{\mathrm{ak}}{=}7$ for InternVL2.5-8B (32 layers). The scaling coefficient $\beta$ in Eq.~\ref{eq:injection} is set to $0.7$ for Qwen2.5-VL-7B, $0.4$ for Qwen3-VL-8B, and $0.7$ for InternVL2.5-8B, and held constant across all benchmarks for each model. Since TAI involves no stochastic components, all results 
are fully deterministic and do not require seed selection 
or repeated runs. All reported accuracies on TempCompass~\citep{liu2024tempcompass}, 
TVBench~\citep{cores2024lost}, AoTBench~\citep{xue2025seeing}, and MVBench~\citep{li2024mvbench} use micro-averaging 
across all models and methods for fair comparison.

\section{Choice of Frame Count}
\label{app:frame-count}

We use $n_{\mathrm{frames}}=16$ throughout this work. Tab.~\ref{tab:frame-interval} reports the median video duration of each benchmark together with the resulting inter-frame interval $\Delta t = \mathrm{duration\ per\ } n_{\mathrm{frames}}$. At $n=16$, the median video yields $\Delta t$ between $0.6$ and $1.3$ seconds, comparable to the duration of the discrete events that these benchmarks probe, such as a person turning, an object falling, or a hand pouring liquid. Doubling to $n=32$ halves these intervals to $0.3$ to $0.6$ seconds, in which case successive frames fall within the same event and add negligible temporal information. This range aligns with the timescale at which humans naturally segment continuous video into discrete events~\citep{kurby2008segmentation}.

A small number of videos are shorter than the typical event timescale itself. Across the four benchmarks, $4.7\%$ of videos have duration below $5$ seconds, with the highest concentration of $11.3\%$ in MVBench, where $16$ frames already approach dense sampling. While both baseline and TAI accuracies are higher at $n=32$ and the improvement due to TAI is also larger (Fig.~\ref{fig:ablation}(a)), this gap reflects redundant within-event sampling rather than additional temporal information. We therefore adopt $n=16$ to align with the event-level granularity of these benchmarks and to ensure fair comparison against reported baselines.

\begin{table}[htbp]
\centering
\caption{Inter-frame sampling intervals at $n_{\mathrm{frames}} \in \{16, 32\}$ on each benchmark, computed at the median video duration. At $n=32$, the interval falls below the duration of typical events probed by these benchmarks (turning, jumping, pouring, etc.), producing redundant within-event sampling.}
\label{tab:frame-interval}
\small
\begin{tabular}{lccc}
\toprule
\textbf{Benchmark} & \textbf{Median duration} & $\Delta t$ \textbf{at} $n{=}16$ & $\Delta t$ \textbf{at} $n{=}32$ \\
\midrule
TempCompass & 10.0\,s & 0.63\,s & 0.31\,s \\
MVBench     & 13.0\,s & 0.81\,s & 0.41\,s \\
TVBench     & 20.0\,s & 1.25\,s & 0.63\,s \\
AoTBench    & 20.0\,s & 1.25\,s & 0.63\,s \\
\bottomrule
\end{tabular}
\end{table}

\section{Baseline Reproduction Details}
\label{sec:reproduction}
We reproduced three baselines under our 16-frame evaluation protocol: TCD~\cite{zhang2024eventhallusion} and DINO-HEAL~\cite{li2025vidhalluc} as training-free methods, and ArrowRL~\cite{xue2025seeing} as a training-based method.

For TCD, we set the number of distorted frames to 4 and swept the weighting coefficient $\alpha \in [0.1, 1.0]$ at $0.1$ intervals, refined to $0.05$ near the optimum, selecting $\alpha{=}0.35$ as the most stable configuration across benchmarks. For DINO-HEAL, we followed the setting reported in the original paper. For ArrowRL, since the method requires additional training and the pretrained weights are publicly available only for Qwen2.5-VL-7B, we used the released checkpoint for reproduction.

\begin{wraptable}{r}{0.34\textwidth}
    \vspace{-7pt}
    \centering
    \small
    \caption{Micro- and macro-averaged AoTBench accuracy.}
    \label{tab:aot-macro}
    \setlength{\tabcolsep}{4pt}
    \begin{tabular}{@{}lcc@{}}
    \toprule
    \textbf{Models} & \textbf{Micro} & \textbf{Macro} \\
    \midrule
    Qwen2.5-VL-7B & 54.5 & 53.1 \\
    +ArrowRL & \textbf{58.2} & \textbf{56.3} \\
    \rowcolor{blue!10} \textbf{+TAI (Ours)} & \underline{57.7} & \underline{55.3} \\
    \midrule
    Qwen3-VL-8B & 56.6 & 54.6 \\
    \rowcolor{blue!10} \textbf{+TAI (Ours)} & \textbf{59.9} & \textbf{57.0} \\
    \midrule
    InternVL2.5-8B & 54.6 & 53.3 \\
    \rowcolor{blue!10} \textbf{+TAI (Ours)} & \textbf{56.6} & \textbf{54.7} \\
    \bottomrule
    \end{tabular}
    \vspace{-10pt}
\end{wraptable}

The ArrowRL scores we report on AoTBench are lower than those in the original paper. Every number in this work comes from our own evaluation under the same hardware and protocol rather than from reported values, since LLM inference results are known to vary across hardware and system configurations~\citep{yuan2025understanding}. The original ArrowRL results are obtained on GH200 GPUs, whereas all our evaluations run on a single A6000 GPU. Tab.~\ref{tab:aotbench-results} and Tab.~\ref{tab:orthogonality} report the micro-average over all AoTBench samples. Tab.~\ref{tab:aot-macro} additionally reports the macro-average over subtasks. TAI improves over the baseline under both conventions on all three models, and the ranking of all methods in Tab.~\ref{tab:aotbench-results} remains identical under macro-averaging. The averaging convention therefore does not affect our conclusions on AoTBench.

We also attempted to reproduce VTD~\cite{qi2025improve} and SEASON~\cite{wu2025seasonmitigatingtemporalhallucination}, both with no official codebase available, but both methods degraded below the unmodified baseline under our 16-frame protocol, as shown in Tab.~\ref{tab:reproduction_failure}. SEASON targets an 8-frame setting, and its temporal homogenization constructs contrastive negatives inside the vision encoder in an architecture-specific manner, making faithful adaptation to other backbones infeasible. VTD was designed for 32-frame inputs, creating a fundamental frame mismatch with our 16-frame protocol. Additionally, the original paper does not define how masked tokens are replaced, leaving a critical implementation detail ambiguous. We therefore excluded both methods from our main comparison.

\begin{table}[htbp]
\centering
\caption{Reproduction of SEASON and VTD on Qwen2.5-VL-7B under our 16-frame protocol, averaged across all TempCompass question formats. Both methods degrade below the baseline, motivating their exclusion from our main comparison.}
\label{tab:reproduction_failure}
\small
\begin{tabular}{lccccc|cc}
\toprule
Method & Act & Attr & Dir & Ord & Spd & AVG & $\Delta$ \\
\midrule
Baseline & 94.9 & 77.2 & 59.1 & 75.8 & 59.2 & 73.4 & --- \\
\midrule
+ SEASON & 93.2 & 72.2 & 53.7 & 71.8 & 58.3 & 70.1 & \textcolor{red}{-3.3} \\
+ VTD & 92.2 & 70.9 & 53.9 & 68.8 & 56.7 & 68.8 & \textcolor{red}{-4.6} \\
\bottomrule
\end{tabular}
\end{table}

\section{Computational Cost Analysis}
\label{app:implementation}
Tab.~\ref{tab:cost-comparison} compares the computational overhead of inference-time methods. All timings are averaged over 100 samples with 3 warmup iterations excluded, measured on the same hardware and input configuration. Contrastive decoding methods such as TCD require up to two full forward passes, while SEASON requires up to three forward passes plus a self-diagnostic step that computes per-layer attention distributions. In our reproduction, the self-diagnostic mechanism requires reading attention weight matrices from intermediate decoder layers, which is incompatible with flash attention. As no official implementation was available, we implemented SEASON with eager attention following the method description in the original paper. VTD similarly requires reading decoder attention weights for its momentum importance scoring, necessitating eager attention as well. DINO-HEAL adds a lightweight DINOv2 forward pass to the standard model forward.The eager attention requirement of SEASON and VTD incurs additional memory overhead from storing full attention matrices. While VTD remains efficient in wall-clock time, SEASON incurs substantial overhead from its three forward passes combined with the self-diagnostic computation, as reflected in Tab.~\ref{tab:cost-comparison}. TAI performs a partial reverse pass up to $L_{\mathrm{src}}$ at ${\sim}0.7\times$ cost with vision feature caching that avoids redundant vision encoding, combined with a full forward pass at $1.0\times$. TAI maintains near-baseline peak memory and is compatible with flash attention, while staying below two full forward passes in wall-clock time, making it competitive in both memory and computational overhead compared to existing inference-time approaches.
\begin{table}[htbp]
  \centering
  \caption{\textbf{Computational overhead comparison} of 
  inference-time methods on Qwen2.5-VL-7B with 16-frame 
  input on a single A6000 GPU. Time averaged over 100 samples.}
  \label{tab:cost-comparison}
  \small
  \begin{tabular}{@{}lccc@{}}
    \toprule
    \textbf{Method} & \textbf{Max. Fwd Passes} & \textbf{Peak Mem (GB)} & \textbf{Time (s/sample)} \\
    \midrule
    Baseline   & $1.0\times$ & 18.66 & $1.14 \pm 0.22$ \\
    DINO-HEAL  & $1.0\times$ + DINOv2 & 19.23 & $1.48 \pm 0.23$ \\
    TCD        & $2.0\times$ & 18.70 & $1.56 \pm 0.20$ \\
    VTD        & $2.0\times$ + distort. & 21.82 & $1.93 \pm 0.23$ \\
    SEASON     & $3.0\times$ + diag. & 24.46 & $5.13 \pm 0.36$ \\
    \rowcolor{blue!10} \textbf{TAI (Ours)} & ${\sim}1.7\times$ & 18.70 & $1.52 \pm 0.12$ \\
    \bottomrule
  \end{tabular}
\end{table}

\section{Details of Fig.~\ref{fig:problem_statement}}
\label{app:fig1-details}

The failure case in Fig.~\ref{fig:problem_statement} is drawn from a real TempCompass binary question evaluated on Qwen2.5-VL-7B. We apply logit lens~\citep{nostalgebraist2020logitlens} at each layer by projecting the last token hidden state through the model's language model head to obtain Yes/No probabilities. The figure displays the first layer L0 and the final 9 layers L19 to L27 of the 28-layer model, where L19 to L22 correspond to approximately 70 to 80\% of the model depth, aligning with the $\hat{\tau}_l$ peak region identified in Sec.~\ref{sec:analysis}. The color gradient indicates the relative probability of the correct answer, with green denoting higher probability for the correct answer and red denoting higher probability for the incorrect answer. For the forward video, the correct answer maintains higher probability than the incorrect answer throughout all displayed layers. For the reversed video, the correct answer emerges at $L_{\mathrm{intermediate}}$ but is progressively overturned by subsequent layers, converging to the same prediction as the forward input by the final layer.

\section{Per-Benchmark Examples of Input-Level $\hat{\tau}$ Variation}
\label{sec:per-benchmark-examples}

We provide qualitative examples from each benchmark showing how $\hat{\tau}_{L_{\mathrm{src}}}$ varies across task categories or subtasks, and across individual inputs within them. The underlying principle is simple. If reversing the input frames does not change the answer, $\hat{\tau}_{L_{\mathrm{src}}}$ is low; if reversal flips the answer, $\hat{\tau}_{L_{\mathrm{src}}}$ is high. All values are measured at $L_{\mathrm{src}}=20$ on Qwen2.5-VL-7B with $n=16$ frames.

\begin{figure}[htbp]
  \centering
  \includegraphics[width=0.8\linewidth]{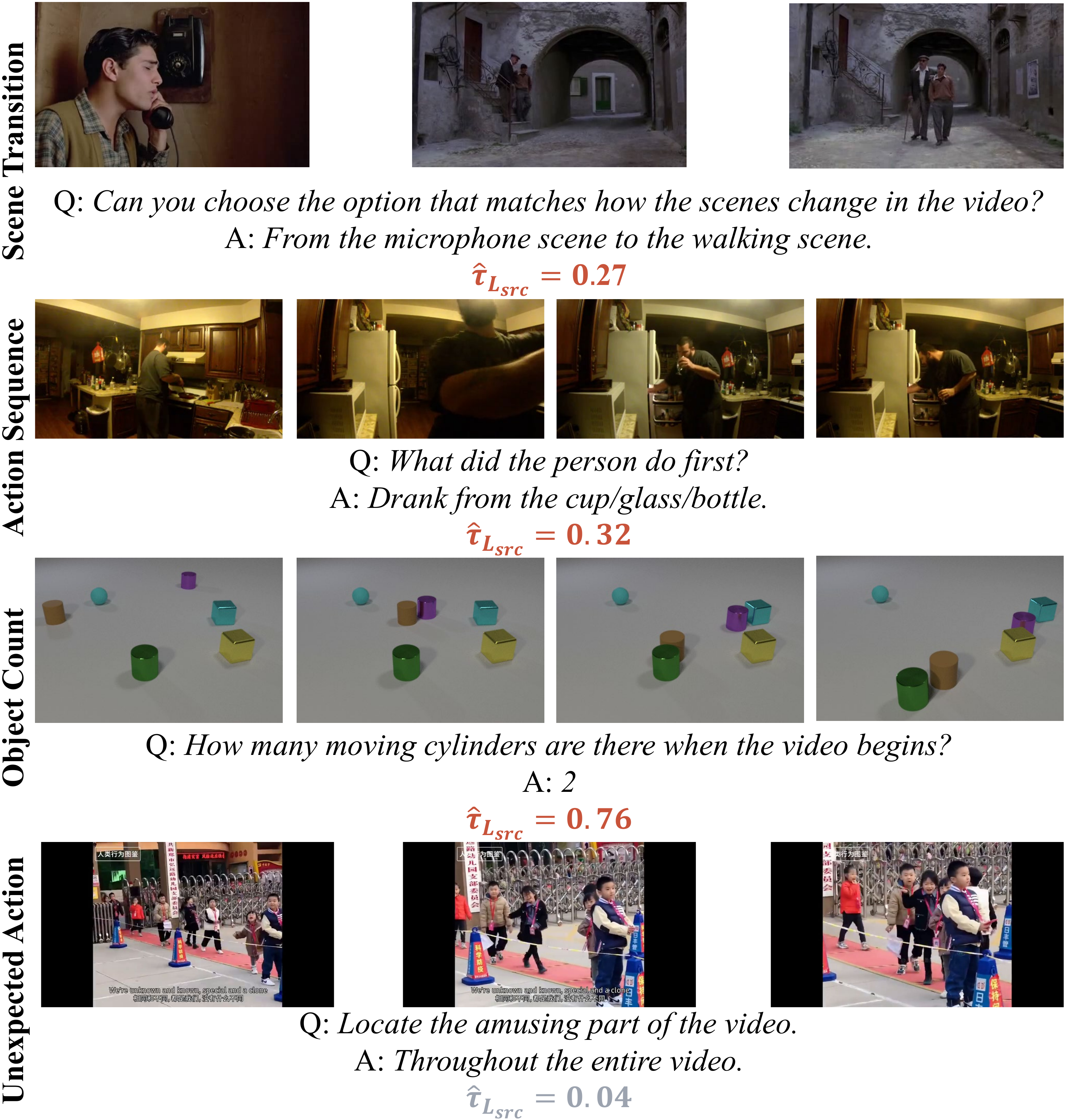}
  \caption{\textbf{TVBench example videos with Scene Transition, Action Sequence, Object Count, and Unexpected Action.} Each row shows the input video, the question prompt, the ground-truth answer, and $\hat{\tau}_{L_{\mathrm{src}}}$.}
  \label{fig:tvbench-examples-1}
\end{figure}

\begin{figure}[htbp]
  \centering
  \includegraphics[width=0.97\linewidth]{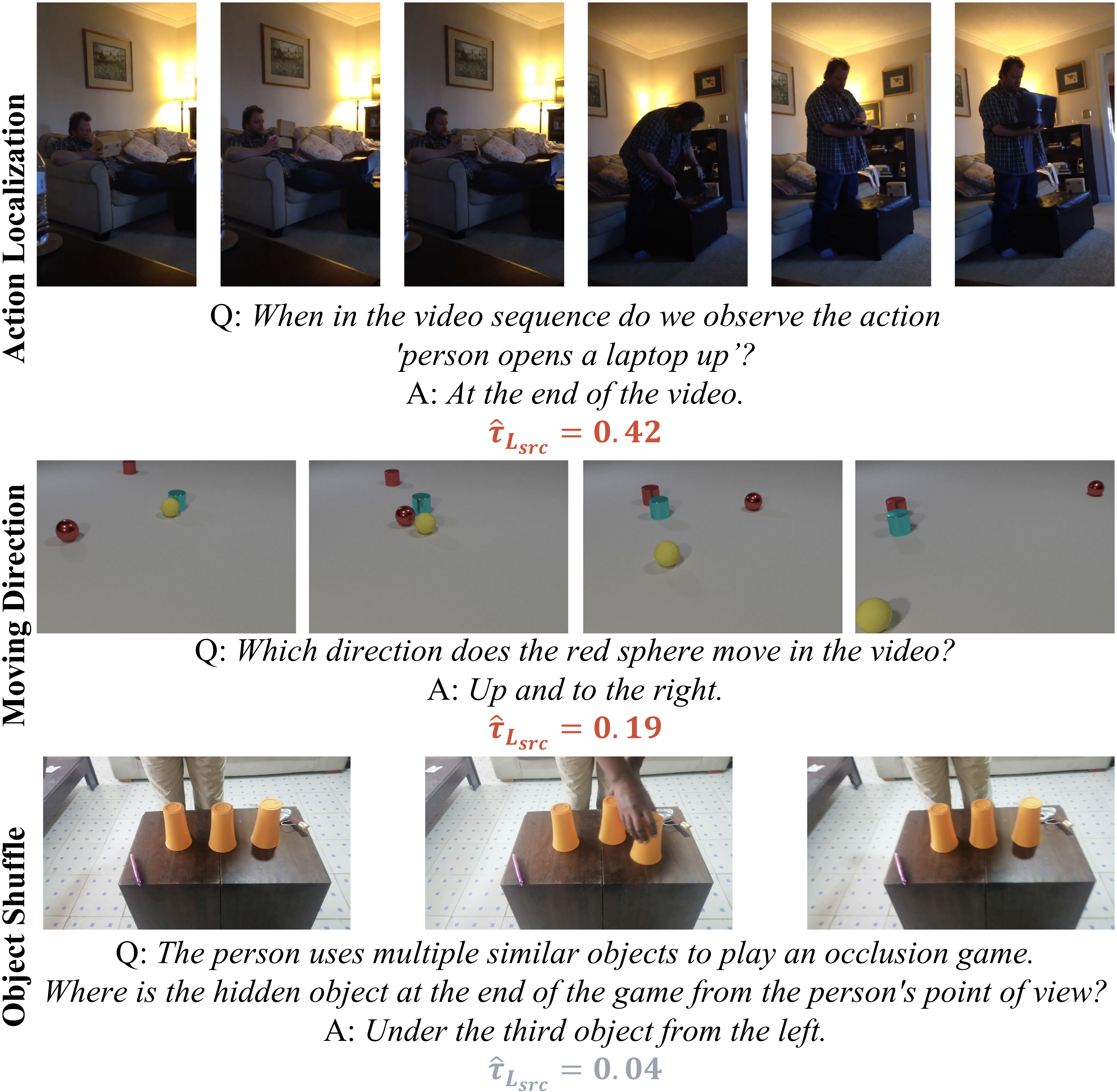}
  \caption{\textbf{TVBench example videos with Action Localization, Moving Direction, and Object Shuffle.} Each row shows the input video, the question prompt, the ground-truth answer, and $\hat{\tau}_{L_{\mathrm{src}}}$.}
  \label{fig:tvbench-examples-2}
\end{figure}

TVBench examples are shown in Figs.~\ref{fig:tvbench-examples-1} and \ref{fig:tvbench-examples-2}. Subtasks on which TAI improves by 2 to 4 points such as Action Localization, Moving Direction, Scene Transition, Action Sequence, and Object Count carry mid-to-high $\hat{\tau}_{L_{\mathrm{src}}}$, while Object Shuffle and Unexpected Action carry low $\hat{\tau}_{L_{\mathrm{src}}}$ and show no TAI gain. The Object Shuffle example is informative. The question concerns the final state of an occlusion game and is order-sensitive in principle, yet the model can answer correctly from the last frame alone, so reversing the frame order barely changes the model's response and $\hat{\tau}_{L_{\mathrm{src}}}$ stays low.

\begin{figure}[htbp]
  \centering
  \includegraphics[width=0.97\linewidth]{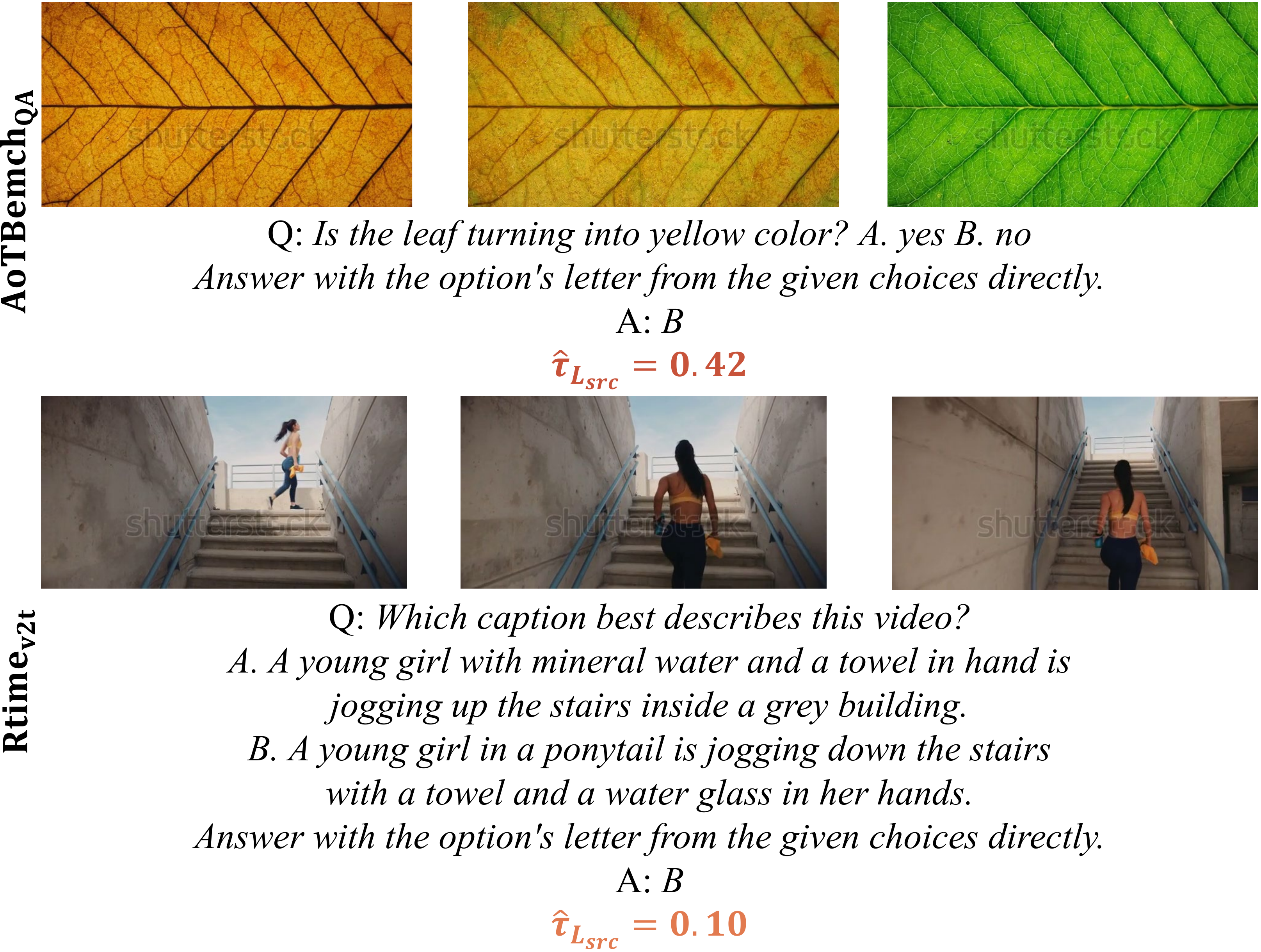}
  \caption{\textbf{AoTBench example videos with AoTBench$_{\mathrm{QA}}$ and Rtime$_{\mathrm{v2t}}$.} Each row shows the input video, the question prompt, the ground-truth answer, and $\hat{\tau}_{L_{\mathrm{src}}}$.}
  \label{fig:aot-examples-1}
\end{figure}

\begin{figure}[htbp]
  \centering
  \includegraphics[width=0.97\linewidth]{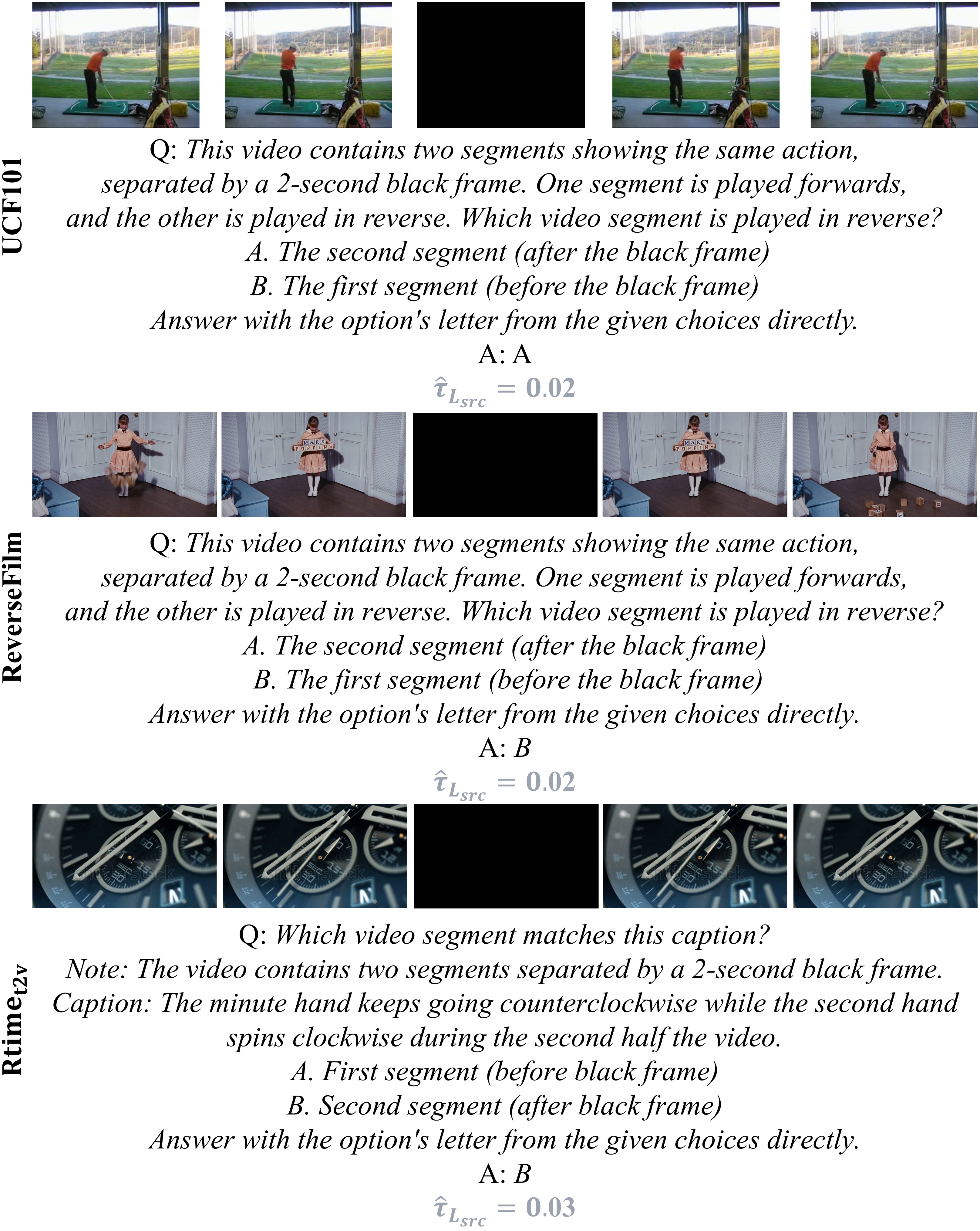}
  \caption{\textbf{AoTBench example videos with UCF101, ReverseFilm, and Rtime$_{\mathrm{t2v}}$.} Each row shows the input video, the question prompt, the ground-truth answer, and $\hat{\tau}_{L_{\mathrm{src}}}$.}
  \label{fig:aot-examples-2}
\end{figure}

AoTBench examples are shown in Figs.~\ref{fig:aot-examples-1} and \ref{fig:aot-examples-2}. The benchmark splits into two structurally different groups. AoTBench$_{\mathrm{QA}}$ and Rtime$_{\mathrm{v2t}}$ use single forward or reversed videos, and the example for AoTBench$_{\mathrm{QA}}$ shows a leaf changing color with a high $\hat{\tau}_{L_{\mathrm{src}}}=0.42$, while the Rtime$_{\mathrm{v2t}}$ example of a person jogging up stairs sits at a moderate $0.10$. UCF101, ReverseFilm, and Rtime$_{\mathrm{t2v}}$ concatenate two segments separated by a 2-second black frame and ask which segment is reversed. Reversing this concatenated input mostly swaps the two halves and preserves nearly all visual content, so $\hat{\tau}_{L_{\mathrm{src}}}$ is uniformly low and TAI gain is negligible.

\begin{figure}[htbp]
  \centering
  \includegraphics[width=0.97\linewidth]{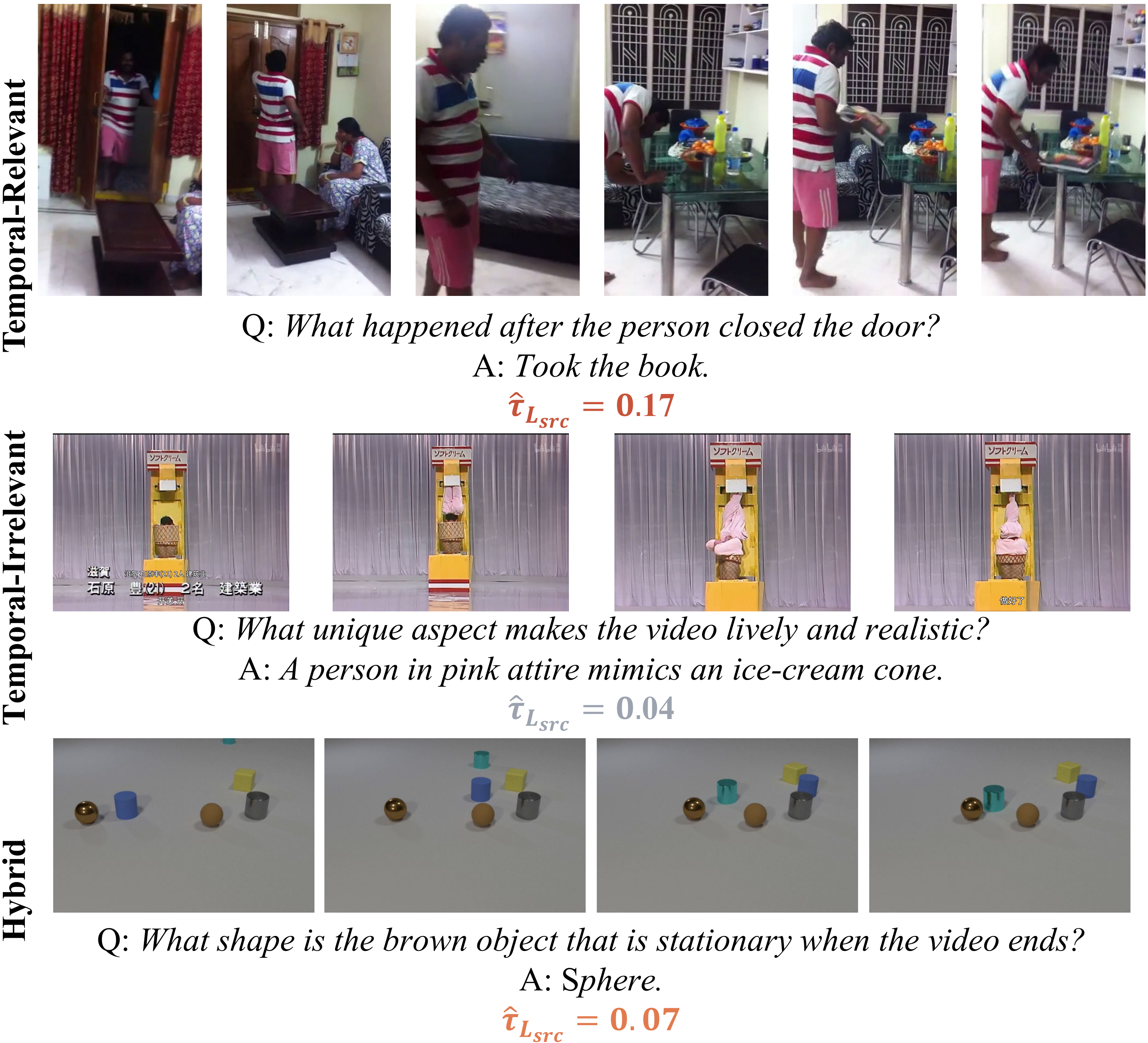}
  \caption{\textbf{MVBench example videos with Temporal-Relevant, Temporal-Irrelevant, and Hybrid groups.} Each row shows the input video, the question prompt, the ground-truth answer, and $\hat{\tau}_{L_{\mathrm{src}}}$.}
  \label{fig:mvbench-examples}
\end{figure}

MVBench examples are shown in Fig.~\ref{fig:mvbench-examples}, with one video from each of the three groups defined in Sec.~\ref{sec:detailed-results}. The Temporal-Relevant example from Action Sequence asks what the person did after closing the door and carries $\hat{\tau}_{L_{\mathrm{src}}}=0.17$. The Temporal-Irrelevant example from Unexpected Action asks what makes the video lively and carries $\hat{\tau}_{L_{\mathrm{src}}}=0.04$. The Hybrid example from Moving Attribute asks the shape of an object stationary at the end and carries $\hat{\tau}_{L_{\mathrm{src}}}=0.07$, falling between the two extremes. The ordering across the three groups mirrors the absolute gain pattern in Tab.~\ref{tab:mvbench-results}.

The qualitative pattern is consistent across the three benchmarks. $\hat{\tau}_{L_{\mathrm{src}}}$ tracks how much the answer depends on the order of the input frames, and the magnitude of TAI's gain on each subtask follows the typical $\hat{\tau}_{L_{\mathrm{src}}}$ of its inputs.

\clearpage

\end{document}